\documentclass{article}

    \PassOptionsToPackage{numbers, compress}{natbib}

\RequirePackage{silence}
\usepackage[preprint]{neurips_2026}

\usepackage[utf8]{inputenc} 
\usepackage[T1]{fontenc}    
\usepackage{hyperref}       
\usepackage{url}            
\usepackage{booktabs}       
\usepackage{amsfonts}       
\usepackage{nicefrac}       
\usepackage{microtype}      
\usepackage{xcolor}         

\usepackage{graphicx}
\usepackage{float}
\usepackage{multirow}

\usepackage{amsmath}
\usepackage{wrapfig}
\title{Causal Object-Centric Models for Planning with Monte Carlo Tree Search}

\author{%
  Rodion Vakhitov \\
  MIRAI\\
  Moscow, Russia \\
  \texttt{vakhitov.r@miriai.org} \\
  \And
  Leonid Ugadiarov \\
  CogAILab \& MIRAI \\
  Moscow, Russia \\
  \AND
  Alexey Skrynnik \\
  CogAILab \& MIRAI \\
  Moscow, Russia \\
  \And
  Aleksandr Panov \\
  CogAILab \& MIRAI \\
  Moscow, Russia\\
}

\begin{document}

\maketitle

\begin{abstract}
We introduce COMET (Causal Object-centric Model for Efficient Tree search), a model-based reinforcement learning algorithm that performs Monte Carlo Tree Search in a slot-structured latent space.
COMET pairs a frozen unsupervised object-centric encoder with a transformer-based world model, in which actions are bound to objects through a novel action–slot fusion mechanism that is used in slot transition prediction.
Policy and value heads use object-causal attention, modulating token interactions by learned per-slot relevance scores so that decision-making concentrates on task-relevant entities.
COMET adds an explicit object-level inductive bias to MuZero-style latent planning.
Across eight visually and dynamically diverse tasks from the Object-Centric Visual RL benchmark, ManiSkill, Robosuite, and VizDoom, COMET achieves a higher mean normalized score during the early stages of training compared to object-centric and monolithic baselines.
\end{abstract}

\section{Introduction}
Humans can reason about the consequences of their actions before acting by mentally simulating past experiences or possible future outcomes~\cite{shiffrin:brain}. Motivated by this ability, world models have been introduced in reinforcement learning (RL) as a way to imitate the environment and improve learning efficiency~\cite{schmidhuber:wm}. In model-based reinforcement learning (MBRL), an agent learns a model of the environment dynamics and uses it to generate imagined experiences, thereby reducing the need for real-world interactions.

MBRL methods have achieved strong performance across a wide range of tasks. Notable examples include the Dreamer family of algorithms~\cite{dreamerv1,dreamerv2,dreamerv3}, which employ latent world models for long-horizon imagination, approaches based on Model Predictive Path Integral (MPPI) control for planning~\cite{tdmpcv1,tdmpcv2}, and methods that integrate Monte-Carlo Tree Search (MCTS)~\cite{coulom2006efficient,schrittwieser2020mastering} with learned models~\cite{schrittwieser2020mastering}.

Despite this progress, learning accurate world models remains difficult in environments that are high-dimensional, non-stationary, and composed of multiple interacting objects.
One of the challenges for visual environments lies in representing observations effectively.
Most existing approaches rely on convolutional neural network (CNN) encoders~\cite{lecun:deeplearning} that produce a single holistic representation of the input image. However, such representations may fail to capture object-level structure and interactions, which are often crucial for decision-making~\cite{santoro:relations}. In complex scenes, small but task-relevant objects, dynamic backgrounds, or many irrelevant entities can significantly degrade agent performance ~\cite{liang2024visarl}.

\begin{wrapfigure}{r}{0.6\textwidth}
\centering
\includegraphics[height=0.55\textwidth]{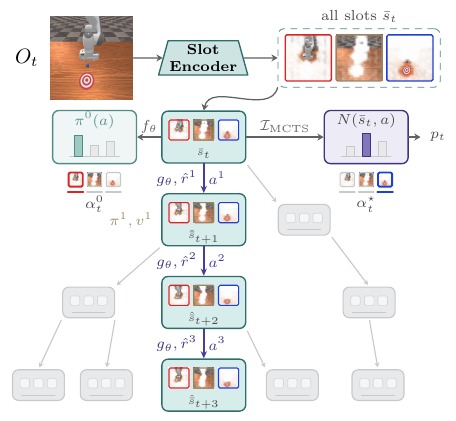}
\caption{Object-centric representations in COMET. Observation $O_t$ is transformed into a set of object representations $\bar{s}_t$, for which causality scores $\bar{\alpha}_t$ are estimated. By focusing on the most relevant objects and their interactions, planning can concentrate on task-relevant elements of the scene during tree search.}\label{fig:va-scheme}
\end{wrapfigure}

Humans, by contrast, perceive the world as composed of discrete entities such as objects~\cite{spelke:coreknowledge}, which enables efficient reasoning and planning. Object-centric RL represents the environment as a set of object-level components, where each component
corresponds to an individual object. When instance segmentation masks are available, object representations can be extracted using CNN encoders, alternatively, supervised segmentation models~\cite{cutie,samv1,samv2} can be used, though they require annotated data. A large body of work instead focuses on unsupervised object-centric representation learning~\cite{cswm,space,locatello2020object,elsayed2022savi++,slate,singh2022simple,dinosaur,videosaur,manasyan2025temporally,ddlp}, which discovers structured representations directly from raw images, making it suitable for reinforcement learning without external supervision.


Object-centric MBRL methods that maintain an object-level world model can explicitly represent object dynamics and interactions, enabling more focused and interpretable decision-making.
Many real-world and simulated environments are inherently object-oriented: scenes consist of multiple objects whose interactions determine the reward.
However, at any given time step, only a small subset of objects typically participates in interactions relevant to the current decision.
For example, in robotic manipulation tasks, a robot often interacts with only one object at a time.
As a result, actions usually affect the state of only a few objects, while the remaining objects are largely irrelevant for the immediate decision.
Motivated by this observation, we hypothesize that explicitly modeling the importance of individual objects for decision-making can improve policy learning.
To this end, we propose COMET, an object-centric MBRL algorithm based on MCTS.
In COMET, the world model maintains disentangled latents for object-centric representations.
The policy and value models use transformer-based architectures~\citep{vaswani2017attention} over these latents, combined with object causal attention mechanisms.
Each network processes object tokens together with a dedicated target token for action or value prediction, while attention is modulated by learned causality scores to emphasize task-relevant objects.

In summary, our main contributions are as follows:
\begin{itemize}
    \item We introduce COMET, an MCTS-based object-centric MBRL algorithm that combines frozen object-level representations with a transformer-based world model for planning in an object-structured latent space.

    \item We propose a novel action-object binding mechanism, where actions are fused with object-centric slots, effectively implementing a learned binding between actions and objects within a unified transformer backbone, enabling object-centric world modeling as well as policy/value prediction.

    \item We evaluate COMET across a diverse set of object-oriented visual control tasks, including object-centric benchmark environments and robotic manipulation tasks, and show that it shows consistent performance across tasks and, on average, achieves higher sample efficiency than both strong monolithic MCTS-based MBRL methods and object-centric RL baselines.
\end{itemize}

\begin{figure}[ht!]
\centering
\includegraphics[width=\textwidth]{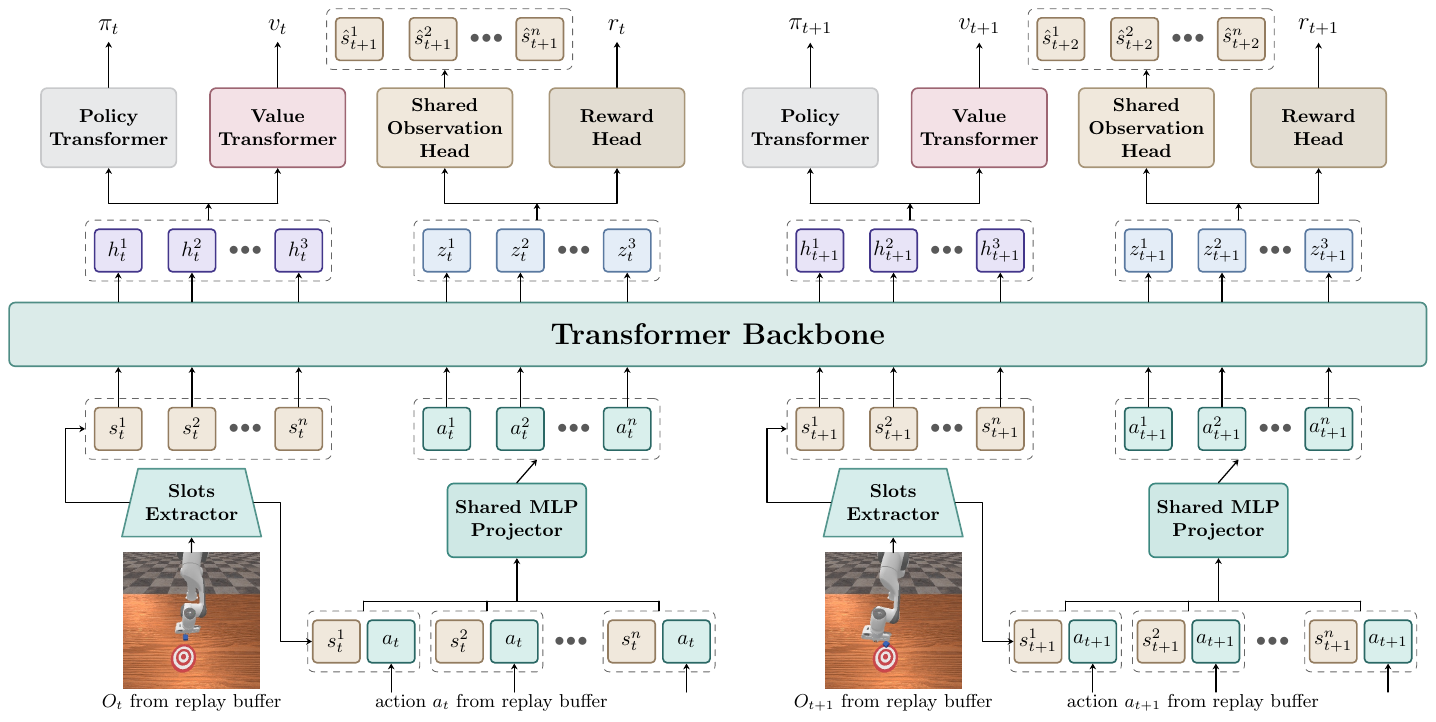}
\caption{Overview of COMET training. A frozen slot extractor maps observations into slots, which are processed by a transformer backbone to produce latent representations $h_t^1, h_t^2, \dots, h_t^n$. These latents, together with a learnable target token, are fed into the policy and value transformers to predict the action distribution or value. Next, an action embedding is concatenated with each slot independently and passed through a shared MLP projector, producing slot-conditioned action embeddings $a_t^1, a_t^2, \dots, a_t^n$. These are processed by a transformer backbone to obtain $z_t^1, z_t^2, \dots, z_t^n$, which are used to predict the next state (next slots) and reward.} \label{fig:main-scheme}
\end{figure}
\section{Related Work}
\subsection{Object-Centric Representation Learning}

A growing line of research focuses on learning structured object-centric representations directly from raw sensory inputs without manual annotations. Instead of encoding a scene into a single global vector, these methods decompose observations into sets of entities that can be processed independently. A key mechanism is Slot Attention~\cite{locatello2020object}, which iteratively assigns a fixed number of latent slots to different parts of the input via competitive cross-attention. Subsequent work extends this idea to sequential data. SAVi~\cite{kipf2022conditional} and SAVi++~\cite{elsayed2022savi++} introduce temporal consistency using motion cues such as optical flow and depth, enabling slots to persist across frames. Other approaches focus on improving reconstruction quality with more expressive generative models. SLATE and STEVE~\cite{singh2022simple} combine discrete latent tokenization (dVAE~\cite{van2017neural}) with transformer-based decoders and Slot Attention-based grouping. In contrast, DINOSAUR~\cite{dinosaur} replaces pixel reconstruction with feature-level objectives using pretrained DINO~\cite{amir2021deep} representations to learn semantically meaningful objects. More recent work Slot Contrast~\cite{manasyan2025temporally} enforces alignment between slots across time by contrasting corresponding object representations, resulting in more robust tracking and reduced slot ambiguity in dynamic scenes.

Not all object-centric models rely on Slot Attention. Deep Latent Particles (DLP)~\cite{daniel2022unsupervised} represent images as low-dimensional particles that decouple spatial position and appearance. In a different direction, Artificial Kuramoto Oscillatory Neurons (AKOrN)~\cite{miyato2024artificial} introduce oscillatory neural dynamics, where synchronized neurons form coherent groups corresponding to objects or parts.

\subsection{Object Centric Reinforcement Learning}
Recent studies incorporate object-centric representations into model-based reinforcement learning to better capture the compositional structure of environments. COBRA~\cite{watters2019cobra} learns a transition model over latent slots obtained from MONet~\cite{burgess2019monet} and combines it with intrinsic motivation to improve data efficiency. FOCUS~\cite{ferraro2023focus} uses an encoder-decoder architecture that segments scenes into object-specific latent variables via learned masks. OC-STORM~\cite{zhang2025objects} employs a spatiotemporal transformer to jointly reason over object-centric and pixel-level representations for dynamics modeling. COBRA is limited by the lack of explicit modeling of object interactions, restricting its ability to capture relational dynamics addressed by our method. In contrast, FOCUS and OC-STORM rely on annotated segmentation masks, limiting their applicability in fully unsupervised settings.

Closer to our setting, STICA~\cite{nishimoto2026object} proposes an object-centric model-based RL framework combining slot-based representations with transformer-based world models and decision modules with object causal attention. SOLD~\cite{mosbach2024sold} learns object-centric latent dynamics directly from pixels without supervision via an action-conditioned slot-based dynamics model and a Slot Aggregation Transformer for policy and value learning. Object-Centric Dreamer~\cite{ugadiarov2025object} (OCDreamer) extends Dreamer by replacing the RSSM with an object-centric RSSM and incorporating GNNs to explicitly model object interactions during prediction and control. Beyond model-based approaches, object-centric representations are also used in model-free RL. OCRL~\cite{yoon2023investigation} integrates a transformer-based object encoder into PPO~\cite{schulman2017proximal}, enabling flexible use of different object-centric features. Similarly, OC-CA and OC-SA~\cite{stanic2023learning} use Slot Attention as a feature extractor and study its generalization across environments.

\section{COMET}
\begin{figure}[ht!] 
\includegraphics[width=\textwidth]{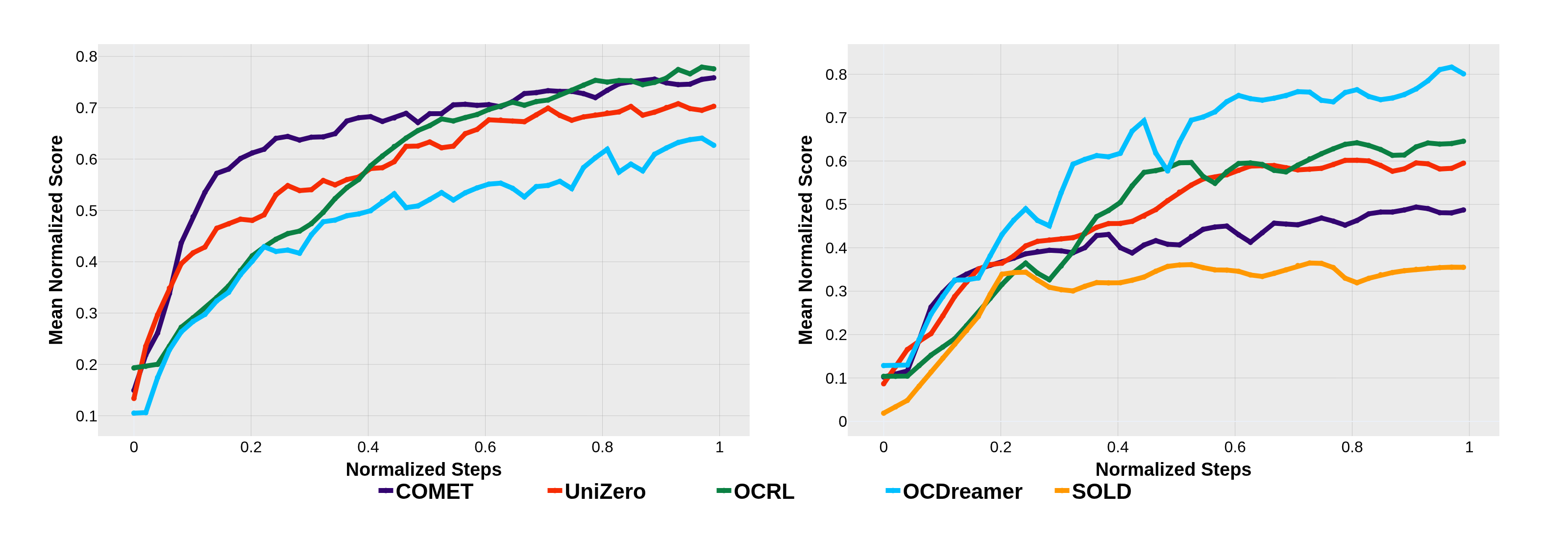}
\vspace{-0.65cm}
\caption{Mean Normalized Score~(\ref{eq:normalized_score}) versus normalized steps.
Normalization parameters are listed in Appendix~\ref{app:normalization}.
\textbf{Left:} normalized score averaged over all considered tasks--Object Goal, Object Interaction, Object Comparison, Property Comparison, Object Reaching, Block Lifting, Cube Pushing, and Defend The Line--for all algorithms except SOLD. 
\textbf{Right:} normalized score averaged over continuous-control tasks compatible with SOLD--Object Reaching, Block Lifting, and Cube Pushing.
} 
\label{fig_score}
\end{figure} 

COMET is an MCTS-based object-centric model-based RL algorithm that performs planning in a slot-structured latent space.
The method combines three components: a frozen object-centric encoder that maps visual observations to object slots, a transformer-based world model that predicts future slots and rewards, and policy/value heads equipped with object-causal attention.
Our implementation builds on the LightZero framework~\cite{niu2023lightzero} and follows the
UniZero training pipeline~\cite{puunizero}.
Unlike UniZero, which operates on monolithic state embeddings, COMET represents each observation as a set of object-centric slots and performs both dynamics prediction and decision-making over these slots.
This design introduces an explicit object-level inductive bias into MuZero-style latent planning, enabling the model to reason over individual entities and their interactions.
As in UniZero, COMET uses the unified transformer backbone implemented using a nanoGPT-based architecture~\citep{karpathy_nanogpt_2023}.

\subsection{Slots Extractor}
\label{subsec:slot_extractor}
The slots extractor maps an image observation $o_t$ into a set of object-centric latent representations. Specifically, it produces an unordered collection of vectors $\bar{s}_t = \{s_t^1, \dots, s_t^n\}$, where $n$ is a fixed hyperparameter conventionally defined as the maximum number of objects in the scene plus one slot for the background.

A key challenge of slot-based architectures, which are trained on static image observations, is that the ordering of slots is not guaranteed to be consistent across time steps due to the permutation-invariant and stochastic nature of slot attention. To mitigate this issue and ensure temporal consistency of object representations within an episode, we initialize the slot representations at time $t+1$ using the slots obtained at time $t$. This encourages stable assignment of slots to underlying objects over time. In contrast, this issue does not arise in video-based object-centric models, where temporal consistency is directly modeled within the architecture. For example, in Slot Contrast, such consistency is handled within the learning pipeline, and no additional slot-initialization mechanism is required, as temporal correspondence is learned end-to-end through the model design.

In our approach, slot extractors are pretrained on observations collected using a random policy and remain frozen during reinforcement learning.  
We experiment with different object-centric representation models depending on the task, including SLATE, DINOSAUR, and Slot Contrast.

\subsection{Object-Centric Token Processing}
UniZero uses a transformer backbone based on the nanoGPT architecture.
It processes sequences of state and action embeddings arranged alternately in a single sequence. In discrete action spaces, actions are represented as learnable embedding vectors, while in continuous action spaces, actions are passed through a two-layer MLP to produce corresponding embeddings.
The transformer processes the sequence in two stages. First, the state embedding $z_t$ is fed into the backbone, producing latent $h_t^z$, which is then passed to the decision head to model policy and value.
Next, the action embedding $a_t$ is processed by the same transformer backbone, yielding latent $h_t^a$, which is passed through the dynamics head to predict the future state $\hat{z}_{t+1}$ and reward $\hat{r}_t$.
UniZero employs standard causal attention masking and learnable absolute positional encodings, with the total sequence length bounded by the context size.

\begin{wrapfigure}{r}{0.35\textwidth}
    \centering
    \includegraphics[width=0.34\textwidth]{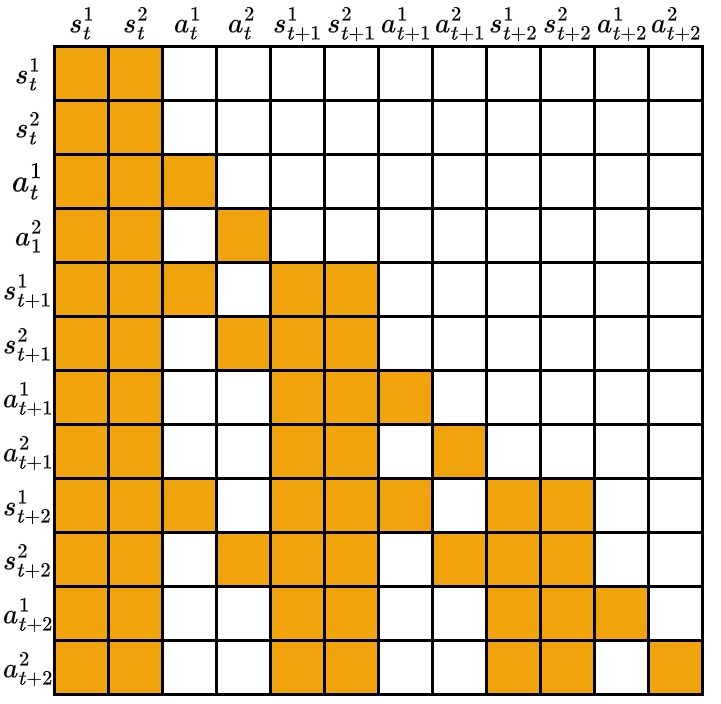}
    \caption{Illustration of the attention mask used in the transformer backbone for a setting with two slots per block across three time steps.}
    \label{fig_attn_mask}
\end{wrapfigure}

We adapt this architecture for object-centric representations.
In object-centric settings, the state is represented as $n$ slots, $\bar{s}_t = \{s_t^1, \dots, s_t^n\}$.
In our approach these slots are fed into the backbone, producing latent representations $\{h_t^1, \dots, h_t^n \}$, which are then passed to the policy and value networks implemented as transformer modules with causal attention, as described in Section~\ref{sec:causal_attn}.
Predicting the next state $\bar{\hat{s}}_{t+1} = \{\hat{s}_t^1, \dots, \hat{s}_t^n\}$ is non-trivial.
Generating $n$ slots from a single action embedding $a_t$ creates a bottleneck, as all object-centric latents must be compressed into a single vector.
In our experiments, architectures using this approach perform poorly.
To address this, we concatenate the action embedding with each slot, ${a_t \oplus s_t^1, \dots, a_t \oplus s_t^n}$, and pass the resulting vectors through a shared MLP projector.
This produces slot-conditioned action embeddings $\bar{a}_t = \{a_t^1, \dots, a_t^n\}$, which are fed into the transformer backbone, yielding latents $\bar{z}_t = \{z_t^1, \dots, z_t^n\}$.
These latents are passed through a shared observation MLP head to predict the next-step object slots $\bar{\hat{s}}_{t+1}$.
For reward prediction, $\bar{z}_t$ are summed into a single vector, which is then processed by a reward head implemented as an MLP.
The input to the transformer is the sequence of $\bar{s}_t$ and $\bar{a}_t$, naturally decomposed into blocks, each corresponding to a single time step.
As in UniZero, each block is augmented with a learnable absolute positional encoding, and the transformer's context size defines the total number of timesteps processed.
In our attention mask, each slot embedding attends to all slots within the same block, all slots from previous blocks, and the action embeddings associated with its position.
Each action embedding attends to itself as well as to all slots in the current and preceding blocks.
The attention mask for the transformer backbone is shown in Figure~\ref{fig_attn_mask}.
We view our slot-conditioned action embeddings as closely related to the mechanism of soft action--object binding~\cite{biza2022binding}, where each slot is influenced by a version of the source action conditioned on the current object state.
The overall training pipeline and architecture are illustrated in Figure~\ref{fig:main-scheme}.

\subsection{Object-Causal Attention}
\label{sec:causal_attn}

The policy and value networks are implemented as a transformer with the modified attention mechanism introduced in STICA.
Alongside the latent representations $h_t^1, \dots, h_t^n$, a learnable target token, specific to the policy or value head, is provided as input; its transformer output is decoded by an MLP to produce the corresponding prediction.
To model causal structure, a causal graph is defined over three groups of objects: the target, causal objects, and non-causal objects:
\begin{equation}
G =
\begin{bmatrix}
1 & 1 & 0 \\
0 & 1 & 0 \\
0 & 0 & 1
\end{bmatrix},
\end{equation}
where $G_{i,j} = 1$ indicates that group $j$ exerts a causal influence on group $i$.
Thus, causal objects influence the target ($G_{1,2}=1$) and one another ($G_{2,2}=1$), while non-causal objects influence only themselves ($G_{3,3}=1$).

Since object causality is not known a priori, a causality score $\alpha_t^k \in [0,1]$ is estimated for each latent $h_t^k$, denoting the probability that the corresponding object is relevant for the policy or value prediction.
\begin{equation}
W_t =
\begin{bmatrix}
1 & 0 & 0 \\
0 & \alpha_t^1 & 1 - \alpha_t^1 \\
\vdots & \vdots & \vdots \\
0 & \alpha_t^n & 1 - \alpha_t^n
\end{bmatrix},
\end{equation}
whose first row corresponds to the target token and remaining rows to the latent object tokens.
The product $W_t G W_t^\top$ lifts $G$ to token-level interactions, encoding the strength of causal influence between every pair of tokens, and is used to modulate scaled dot-product attention:
\begin{equation}
\mathrm{CA}_t = \mathrm{Norm}\!\left(
\mathrm{softmax}\!\left( \frac{Q_t K_t^\top}{\sqrt{d}} \right)
\odot W_t G W_t^\top
\right) V_t,
\end{equation}
where $Q_t$, $K_t$, $V_t$ are the query, key, and value matrices, $d$ is the key dimensionality, $\odot$ is element-wise multiplication, and $\mathrm{Norm}(\cdot)$ denotes row-wise normalization.
The mechanism thereby concentrates attention on objects that directly influence the target and suppresses irrelevant ones.
Although we follow the terminology of STICA and refer to these quantities as causality scores,
they should be interpreted as learned object-relevance weights rather than independently
identified causal effects.

\subsection{Policy and World Model Learning}
\label{sec:learning}
MuZero-like methods learn a latent model for planning with MCTS rather than
using the true environment dynamics \cite{schrittwieser2020mastering}.
The model consists of a representation function, a dynamics function, and a prediction function.
The representation function encodes the observation history into a root latent state $x_t = h_\theta(o_{\leq t}, a_{<t})$.
The dynamics function predicts imagined transitions and rewards, $\hat{r}_t, x_{t+1} = g_\theta(x_t, a_t)$, and the prediction function outputs a policy prior and value estimate, $\pi_t, \hat{v}_t = f_\theta(x_t)$.
MCTS is then performed entirely in latent space. The learned dynamics expands candidate future trajectories, while the prediction function evaluates latent states and provides action priors.
After a fixed number of simulations, the visit counts $N(x_t, a_t)$ at the root are normalized to produce an improved policy target $p_t$

\begin{equation}
p_t =
\frac{N(x_t, a_t)^{1/\mathbb{T}}}{\sum_{b_t} N(x_t, b_t)^{1/\mathbb{T}}},
\end{equation}

where $\mathbb{T}$ denotes the temperature, which modulates the extent of exploration, the visit count $N(x_t, a_t)$ denotes the number of times action $a_t$ was selected at the root latent state $x_t$ during MCTS.
The denominator $\sum_{b_t} N(x_t, b_t)^{1/\mathbb{T}}$ sums over all possible actions $b_t$ from the same state $x_t$ for discrete action spaces, or over sampled candidate actions in continuous-control tasks, where $N(x_t, b_t)$ is the visit count for action $b_t$.
The model is trained end-to-end by unrolling the dynamics for $K$ steps and optimizing policy, value, and reward prediction losses.
This framework forms the basis of several MuZero-style algorithms
\cite{schrittwieser2020mastering,sampled-muzero,efficientzerov1,efficientzerov2}.

While standard MuZero-like methods represent the planning state as a monolithic latent vector or feature map, COMET represents it as a set of object-centric slots, enabling planning and prediction over slot-structured latent space.
The joint optimization objective for COMET can be written as
\begin{equation}
\begin{split}
\mathcal{L}_{\text{COMET}}(\theta)
=
\mathbb{E}_{(o_t, a_t, r_t, o_{t+1}, p_t) \sim \mathcal{B}}
\Big(
\sum_{t=0}^{H-1}
\Big(
\beta_s \frac{1}{n} \sum_{i=1}^{n} \|\hat{s}_{t+1}^{i} - s_{t+1}^{i}\|_2^2 \\
+ \beta_r \,\mathrm{CE}(\hat{r}_t, r_t)
+ \beta_p \,\mathrm{CE}(\pi_t, p_t)
+ \beta_v \,\mathrm{CE}(\hat{v}_t, v_t)
\Big)
\Big),
\end{split}
\end{equation}

where $\mathcal{B}$ is a replay buffer that stores trajectories $\{o_t, a_t, r_t, o_{t+1}, p_t\}$. $H$ denotes the training context length, which corresponds to the rollout length used during training. In COMET, $H$ matches the context window of the transformer backbone. The coefficients $\beta_s, \beta_r, \beta_p, \beta_v$ 
are constant coefficients used to balance different loss terms, corresponding to next-state prediction, reward prediction, policy prediction, and value prediction, respectively. 
$\mathrm{CE}$ denotes the cross-entropy loss function.
Following UniZero, we formulate reward and value prediction as a discrete regression problem in a log-transformed space, optimized by minimizing cross-entropy using $v_t$ and $r_t$ as soft targets.
$\hat{s}_{t+1}^{i}$ denotes the predicted representation of the $i$-th slot, while $s_{t+1}^{i}$ denotes the corresponding ground-truth slot obtained from a frozen pretrained object-centric encoder. 
$v_t$ denotes the bootstrapped $n$-step TD target, and $r_t$ denotes the target reward.

\section{Experimental Setup}
\subsection{Environments}

\begin{figure}[ht!] 
\includegraphics[width=\textwidth]{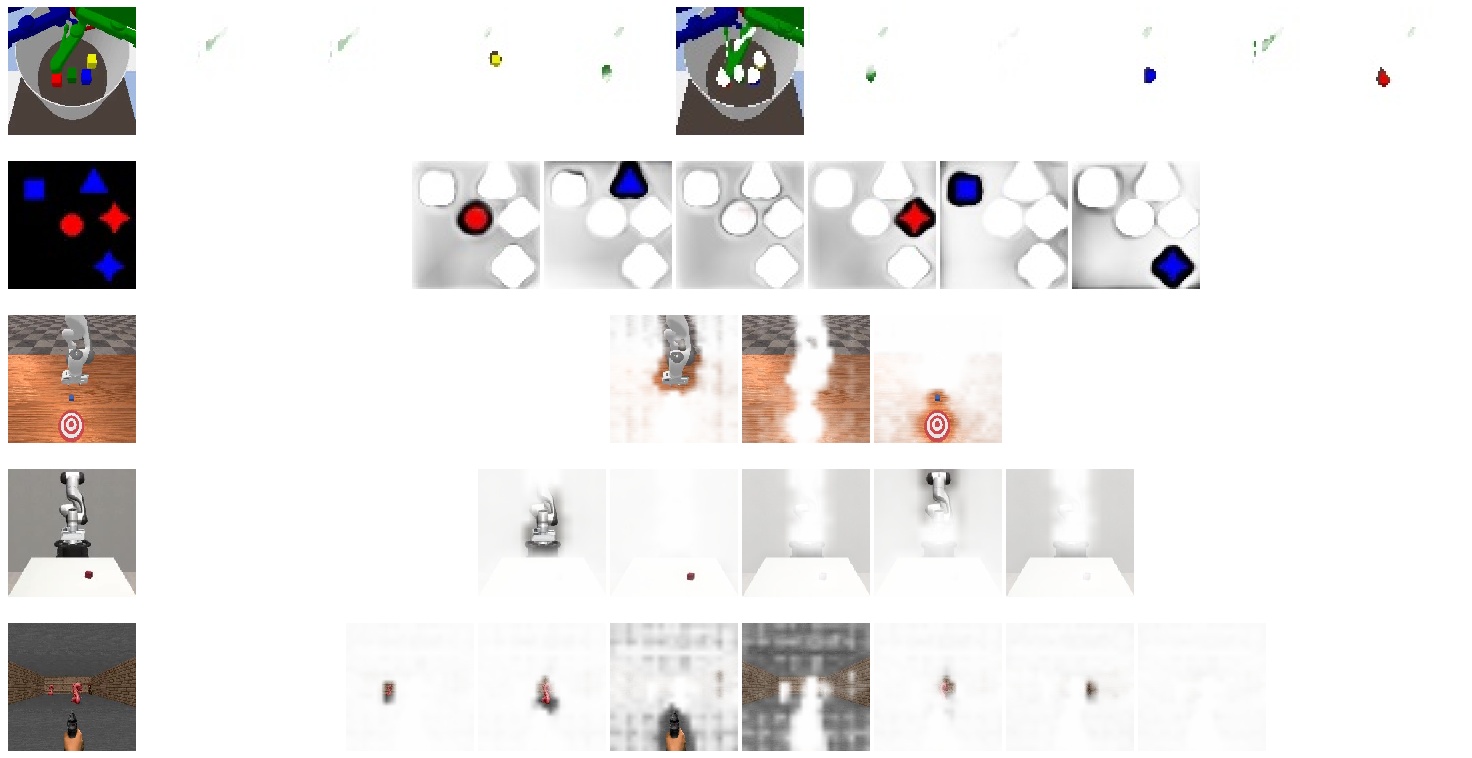}
\vspace{-0.65cm}
\caption{Visualization of observations and slot-wise attention maps across environments. In each row, the real observation is followed by attention maps for each slot produced by the corresponding model. From top to bottom: SLATE in Object Reaching Task, SLATE in an Object Goal Task, Slot Contrast in Cube Pushing Task, DINOSAUR in Block Lifting Task, and Slot Contrast in Defend The Line Task.} 
\label{fig_envs}
\end{figure}

We evaluate our approach on the Object-Centric Visual RL benchmark~\cite{yoon2023investigation}, which includes object-centric environments designed to test perception, interaction, and relational reasoning. The suite consists of Object Goal, Object Interaction, Object Comparison, Property Comparison, and Object Reaching tasks, featuring target objects with distractors and requiring different forms of goal-directed behavior. Across tasks, the agent must identify relevant objects, reason about their properties or relationships, and act under sparse rewards with either discrete or continuous action spaces. We further extend the evaluation to manipulation and control tasks from ManiSkill, Robosuite, and VizDoom. Figure~\ref{fig_envs} shows examples of observations from the considered environments.

In the ManiSkill framework, we use the Cube Pushing task, where a cube is placed on a tabletop and its initial position is randomly sampled within a small region in front of the agent. The goal is to push the cube into a target region at a fixed offset from its initial position, indicated by a visual marker. The reward is dense and shaped, with pose-based components weighted by $\texttt{pose\_reward\_coef}=0.01$ and $\texttt{place\_reward\_coef}=0.1$. Episodes are limited to 50 steps.

In Robosuite, we evaluate the Block Lifting task, where a single Panda arm operates in a tabletop environment. A cube is placed on the table at a fixed position, while the robot’s initial configuration is randomized at each episode. The objective is to grasp and lift the cube above a predefined height threshold. The task uses a dense reward that encourages gradual progress toward successful lifting, requiring stable grasping and vertical manipulation. Episodes are limited to 125 steps.

Finally, in VizDoom we use the Defend the Line scenario, where the agent is placed on one side of a rectangular map, while melee and ranged monsters spawn on the opposite side and continuously move toward it. Monsters are eliminated with a single shot and respawn after a delay; over time they deal increasing damage. The agent has limited ammunition, and the episode ends when the agent dies. The reward is +1 for killing a monster and -1 for death.

\begin{figure}[ht!] 
\includegraphics[width=\textwidth]{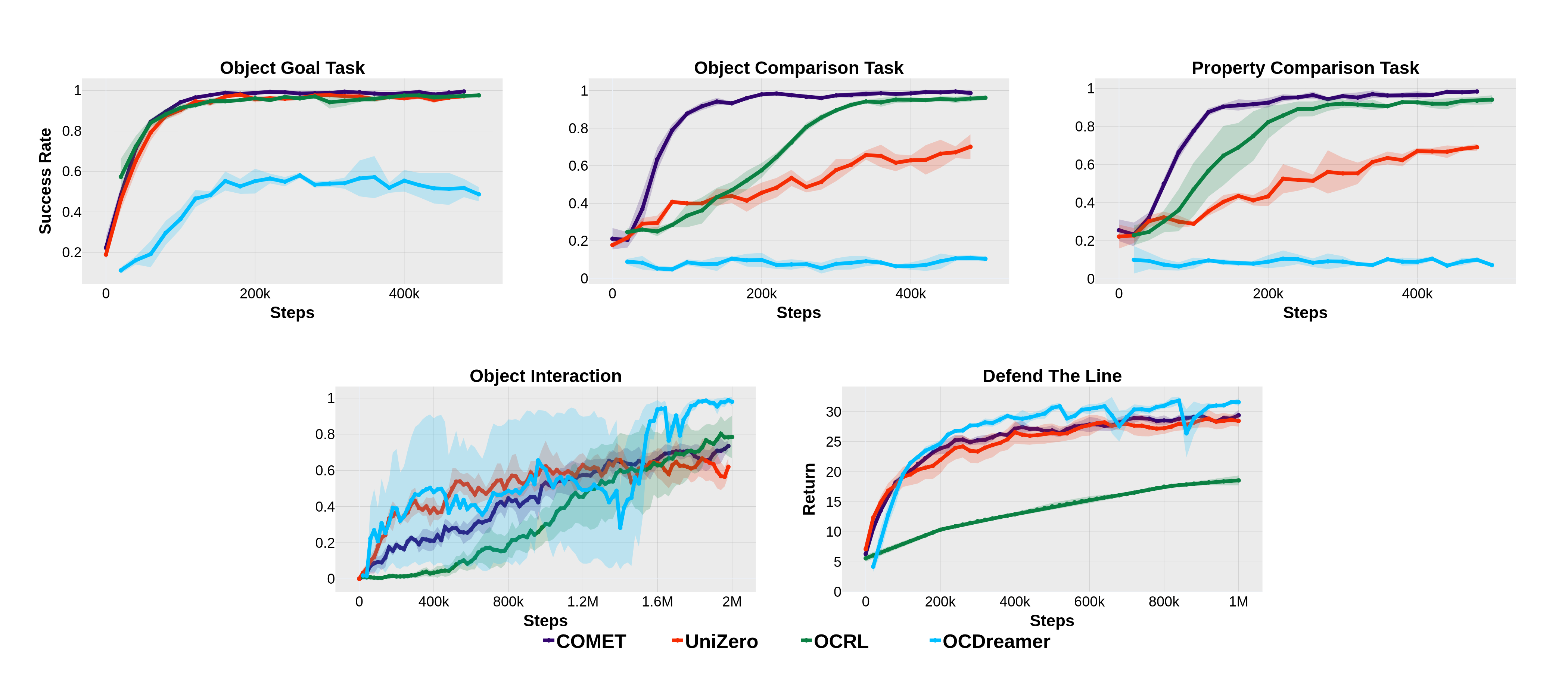}
\vspace{-0.65cm}
\caption{Success rate averaged over 30 episodes and three seeds for COMET and baselines  in tasks with discrete action space. Shaded areas indicate standard deviation. Exponential smoothing with coefficient 0.5 is applied.} 
\label{fig3}
\end{figure}


\begin{figure}[ht!] 
\includegraphics[width=\textwidth]{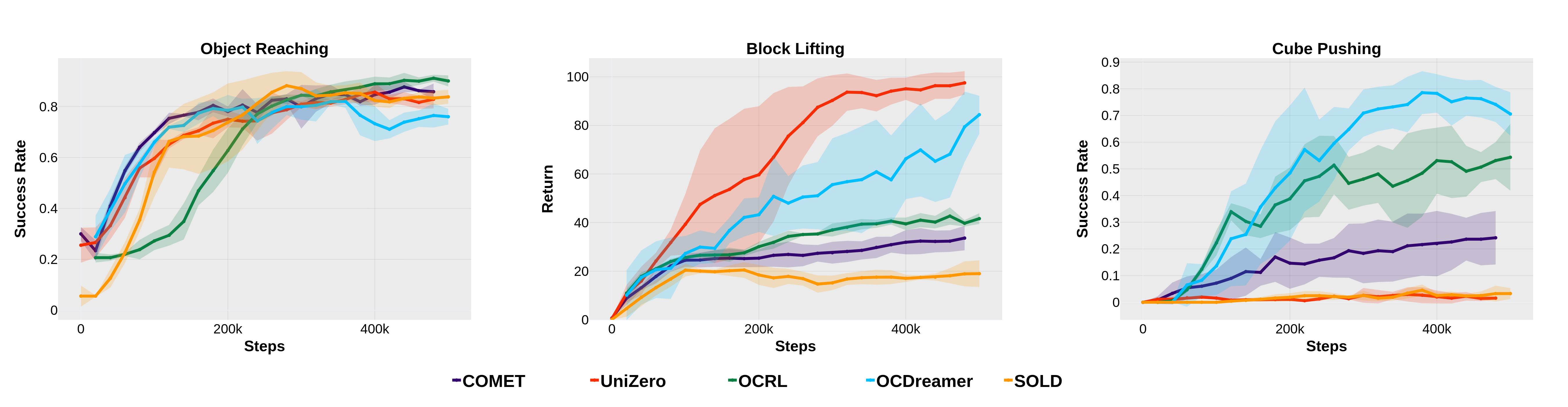}
\vspace{-0.8cm}
\caption{Success rate averaged over 30 episodes and three seeds for COMET and baselines in tasks with continuous action space. Shaded areas indicate standard deviation. Exponential smoothing with coefficient 0.5 is applied.} 
\label{fig4}
\end{figure} 

\subsection{Mean Normalized Score}
\label{sec:score}
The performance and sample efficiency of object-centric RL depend on the quality of learned representations, which are strongly affected by visual complexity. Although agents can sometimes compensate for imperfect representations, this usually reduces sample efficiency and final performance, and different architectures vary in their robustness across environments. Therefore, a key goal is to design object-centric RL agents that maintain stable performance across visually diverse tasks under limited interaction budgets.

Our task suite spans environments with varying visual complexity, ranging from relatively simple settings (e.g., 2D shapes on a monochrome background) to more challenging ones (e.g., ManiSkill, Robosuite, and VizDoom).
To quantify performance under a fixed interaction budget, we compute the average performance across environments.
To this end, we introduce a normalized score that aggregates an agent's performance across a set of visually and dynamically diverse tasks into a single interpretable metric.  
Because evaluation metrics (e.g., cumulative reward and success rate) differ in scale and nature across tasks, raw scores must be normalized prior to aggregation.
We normalize each agent's performance relative to the best-performing method in each environment.
Let $\mathcal{E}$ denote the set of environments, $\mathcal{T}_e$ the time-step budget for environment $e \in \mathcal{E}$, $\mathcal{M}$ the set of RL agents (algorithms) we evaluate, $\tau = t / \mathcal{T}_e \le 1$ the normalized time step, and $F(m, e; t)$ the evaluation metric (success rate or cumulative reward) achieved by method $m \in \mathcal{M}$ on environment $e$ after $t$ training steps.  
We define the normalized score $\mathcal{S}(m,e; \tau)$ for method $m$ in environment $e$, and the mean normalized score $\hat{\mathcal{S}}(m; \tau)$ that aggregates performance across environments:
\begin{equation} \label{eq:normalized_score}
    \mathcal{S}(m, e; \tau) = \dfrac{F (m, e; t)}{\max_{m', t'} F(m', e; t')} \le 1, \quad 
    \hat{\mathcal{S}}(m; \tau) = \dfrac{1}{|\mathcal{E}|} \sum_{e \in \mathcal{E}} \mathcal{S}(m, e; \tau).
\end{equation}

\section{Experiments}
We compare COMET against a model-free, object-centric PPO baseline OCRL~\cite{yoon2023investigation} that uses a transformer encoder to pool object-centric representations.
As object-centric MBRL baselines, we use OCDreamer and SOLD.
OCDreamer is agnostic to the action space, whereas SOLD is implemented only for continuous action spaces.
We experimented with a discrete-action variant by replacing the continuous actor with a categorical policy and optimizing imagined rollouts with a score-function estimator, but this variant failed to learn reliably in preliminary experiments.
We therefore restrict SOLD comparisons to continuous-action tasks, where the original algorithm is directly applicable.
For SOLD, we use the SAVi encoder from the original implementation; examples of its attention maps on continuous tasks are provided in Appendix~\ref{app:savi}.
For OCRL, OCDreamer, and COMET, we use SLATE on the Object Goal, Object Interaction, Property Comparison, Object Property, and Object Reaching tasks; DINOSAUR on Block Lifting; and Slot Contrast on Cube Pushing and Defend the Line.
Examples of attention maps produced by these encoders are shown in Figure~\ref{fig_envs}.
Pre-trained SLATE is used as described in Appendix~\ref{app:slate}, while all other encoders are trained on the collected data (Appendix~\ref{app:datasets}) and kept frozen thereafter.
Hyperparameters for SLATE, Slot Contrast, and DINOSAUR are provided in Appendices~\ref{app:slate}, \ref{app:slotcontrast}, and~\ref{app:dinosaur}, respectively.
As a monolithic MBRL baseline, we use UniZero.
For all baselines, we adopt the original hyperparameters specified in their respective publications and official repositories.
Hyperparameters for COMET are provided in Appendix~\ref{app:comet}.
For the Cube Pushing task, we reduce \texttt{discount\_factor} to $0.925$ for both COMET and UniZero.
True and predicted trajectory rollouts for all tasks are presented in Appendix~\ref{app:rollout}, and causality-score visualizations for the policy and value transformers across all tasks are presented in Appendix~\ref{app:probs}.

Figures~\ref{fig3} and~\ref{fig4} show the training curves of success rate and cumulative reward for COMET and the baselines across all tasks.
The set of top-performing algorithms changes across tasks.
COMET achieves faster convergence and higher final performance than the baselines on Object Comparison and Object Property, but does not outperform the competing methods on Block Lifting.
To account for variability in task difficulty and performance scales, we normalize the results per task for each algorithm as described in Equation~\ref{eq:normalized_score}.
The resulting normalized score, shown in Figure~\ref{fig_score}, demonstrates that COMET achieves higher mean normalized score across a visually diverse set of tasks. 

In visually simple tasks, such as Object Goal, Object Comparison, and Property Comparison, COMET leverages strong object-centric representations and a structure with a single target object and multiple distractors.
Using object causal attention, which benefits from high-quality representations, COMET accurately identifies task-relevant objects by assigning them higher causality scores, as illustrated in Appendix~\ref{app:probs}, enabling it to outperform the baselines.
In more dynamically complex tasks, such as Defend the Line and Object Interaction, COMET achieves performance comparable to the baselines.
We attribute this behavior to task-specific factors: in Defend the Line, most objects are relevant for prediction, while in Object Interaction, the agent must rely on object-pushing mechanics, so the goal is achieved indirectly through deeper causal chains than those captured by our causal object attention mechanism.
For tasks that are challenging in both visual complexity and control, such as Block Lifting and Cube Pushing, COMET achieves moderate performance.
This is likely due to limitations in the object-centric representation model, which sometimes merges the cube and the background into a single slot.
OCDreamer demonstrates better performance on these tasks, indicating that it is more robust to imperfect representations in such settings.

\section{Limitations \& Future Work}
Despite advances in unsupervised object-centric representation learning, current methods still struggle to reliably segment complex, cluttered real-world scenes, especially under occlusion or ambiguous boundaries, limiting their applicability in unconstrained settings.
Additionally, transformer-based approaches scale poorly due to quadratic self-attention costs with increasing object slots, restricting scalability in multi-object scenes.
Future work will focus on extending these methods to realistic, open-ended environments, such as household tasks, and on improving causal attention mechanisms that dynamically prioritize relevant objects.
Enhancing these mechanisms for not only policy and value estimation but also transition dynamics could improve efficiency, relational reasoning, and scalability in complex, multi-object settings.

\section{Conclusion}
In this work, we introduced COMET, an MCTS-based object-centric model-based reinforcement learning method that combines structured object-centric representations with a transformer-based world model.
COMET employs an action-slot binding mechanism that fuses object-centric slots with actions, enabling transition modeling for slots within a unified transformer backbone.
By leveraging a causal attention mechanism, the policy and value models focus on task-relevant object representations, improving both the effectiveness and interpretability of decision-making.
Experimental results across visually diverse discrete and continuous control tasks show that COMET achieves a higher mean normalized score during the early stages of training compared to object-centric and monolithic baselines.

\bibliographystyle{plainnat}
\bibliography{bib}

\clearpage
\appendix

\section{Mean Normalized Score} \label{app:normalization}
\begin{table}[h]
\centering
\scriptsize
\caption{Normalization Parameters. Notation is consistent with Equation~(\ref{eq:normalized_score}).}
\label{tab:nomralization_parameters}
\begin{tabular}{llcc}
\toprule
Task & Metric, $m$ & Normalization Value, $\max_{m', t'} F(m', e; t')$ & Timestep Budget, $\mathcal{T}_e$ \\
\midrule
Object Goal & Success Rate & 1 & 500 k \\
Object Interaction & Success Rate & 1 & 2 M \\
Object Comparison & Success Rate & 1 & 500 k \\
Property Comparison & Success Rate & 1 & 500 k \\
Object Reaching & Success Rate & 1 & 500 k \\
Block Lifting & Cumulative Reward & 104.6 & 500 k \\
Cube Pushing & Success Rate & 1 & 500 k \\
Defend The Line & Cumulative Reward & 33.6 & 1 M \\
\bottomrule
\end{tabular}
\end{table}

\section{Datasets} \label{app:datasets}
We collect images to train object-centric representation models using a uniform random policy.
\begin{table}[h]
\centering
\caption{Parameters of the collected datasets.}
\label{tab:datasets_parameters}
\begin{tabular}{lcc}
\toprule
Task & Dataset Size & Source Image Resolution \\
\midrule
Block Lifting & 300 k & $256 \times 256$ \\
Cube Pushing & 300 k & $224 \times 224$ \\
Defend The Line & 500 k & $336 \times 336$ \\
\bottomrule
\end{tabular}
\end{table}

\clearpage

\section{SLATE`s Hyperparameters} \label{app:slate}
For Object Goal, Object Interaction, Object Comparison, Property Comparison, and Object Reaching tasks we used pre-trained SLATE models from the OCRL official repository \url{https://github.com/jsikyoon/OCRL}.
\begin{table}[h]
\centering
\caption{Hyperparameters for the SLATE}
\label{tab:causalworld_hyperparameters}
\begin{tabular}{lll}
\toprule
\multirow{12}{*}{Learning}
 & Training dataset size & 1000000 \\
 & Temp. cooldown & 1.0 to 0.1 \\
 & Temp. cooldown steps & 30000 \\
 & LR for DVAE & 0.0003 \\
 & LR for CNN Encoder & 0.0001 \\
 & LR for Transformer Decoder & 0.0003 \\
 & LR warm-up steps & 30000 \\
 & LR half time & 250000 \\
 & Dropout & 0.1 \\
 & Clip & 0.05 \\
 & Batch size & 32 \\
 & Epochs & 100 \\
\midrule
\multirow{1}{*}{DVAE}
 & Vocabulary size & 4096 \\
\midrule
\multirow{1}{*}{CNN Encoder}\
 & Hidden size & 64 \\
\midrule
\multirow{4}{*}{Slot Attention}
 & Iterations & 3 \\
 & Slot heads & 1 \\
 & Slot dim. & 192 \\
 & MLP hidden dim. & 192 \\
\midrule
\multirow{3}{*}{Transformer Decoder}
 & Layers & 4 \\
 & Heads & 4 \\
 & Hidden dim. & 192 \\
\bottomrule
\end{tabular}
\end{table}

\clearpage

\section{DINOSAUR`s Hyperparameters} \label{app:dinosaur}
\begin{table}[h]
\centering
\caption{Hyperparameters for DINOSAUR}
\label{tab:dinosaur_hyperparameters}
\begin{tabular}{lll}
\toprule
\multirow{13}{*}{Learning} 
 & Training dataset size & 300000 \\
 & Training steps & 500000 \\
 & Batch size & 64 \\
 & LR warm-up steps & 10000 \\
 & Peak LR & 0.0004 \\
 & Exp. decay half-life & 100000 \\
 & ViT Architecture & ViT-B \\
 & Feature dim. & 768 \\
 & Patch size & 8 \\
 & Gradient norm clipping & 1.0 \\
 & Image/Crop size & 224 \\
 & Cropping strategy & Full \\
 & Tokens & 784 \\
\midrule
\multirow{3}{*}{Decoder}
 & Type & MLP \\
 & Layers & 4 \\
 & MLP hidden dim. & 1024 \\
\midrule
\multirow{4}{*}{Slot Attention}
 & Iterations & 3 \\
 & Slots & 5 \\
 & Slot dim. & 64 \\
 & MLP hidden dim. & 512 \\
\bottomrule
\end{tabular}
\end{table}

\vspace{1em}

\section{Slot Contrast`s Hyperparameters} \label{app:slotcontrast}
\begin{table}[h]
\centering
\caption{Hyperparameters for Slot Contrast}
\label{tab:slotcontrast_hyperparameters}
\begin{tabular}{lll}
\toprule
\multirow{13}{*}{Learning} 
 & Training steps & 100000 \\
 & Batch size & 64 \\
 & Training segment length & 4 \\
 & LR warm-up steps & 2500 \\
 & Optimizer & Adam \\
 & Peak LR & 0.0004 \\
 & Exp. decay half-life & 100000 \\
 & ViT Architecture & DINOv2 Small \\
 & Initialization & FixedLearnedInit \\
 & Patch size & 14 \\
 & Feature dim. & 384 \\
 & Gradient norm clipping & 0.05 \\
 & Image/Crop size & 336 \\
 & Cropping strategy & Full \\
 & Image tokens & 576 \\
\midrule
\multirow{3}{*}{Predictor}
 & Type & Transformer \\
 & Layers & 1 \\
 & Heads & 4 \\
\midrule
\multirow{1}{*}{Decoder}
 & Type & MLP \\
\midrule
\multirow{2}{*}{Slot Attention}
 & Iterations (first / other frames) & 3 / 2 \\
 & Slot dim. & 64 \\
\midrule
\multirow{1}{*}{Loss Parameters}
 & Slot-slot contrastive loss & disabled \\
\bottomrule
\end{tabular}
\end{table}

\clearpage

\section{COMET`s Hyperparameters} \label{app:comet}

\begin{table}[h]
\centering
\caption{Hyperparameters for COMET}
\label{tab:unizero_hyperparameters}
\resizebox{\textwidth}{!}{
\begin{tabular}{lll}
\toprule

\multirow{10}{*}{Planning}
 & Number of MCTS Simulations (sim) & 50 \\
 & Number of Sampled Actions (K) & 20 (Continuous tasks only) \\
 & Inference Context Length & 10 \\
 & Temperature & 0.25 \\
 & Dirichlet Noise ($\alpha$) & 0.3 \\
 & Dirichlet Noise Weight & 0.25 \\
 & Coefficient $c_1$ & 1.25 \\
 & Coefficient $c_2$ & 19652 \\
\midrule

\multirow{6}{*}{Environment and Replay Buffer}
 & Replay Buffer Capacity & 1,000,000 \\
 & Sampling Strategy & Uniform \\
 & Reward Clipping & True (Discrete only) \\
 & Data Augmentation & False \\
 & Game Segment Length & 400 (Discrete); 100 (Continuous) \\
\midrule

\multirow{7}{*}{Architecture}
 & Number of Backbone Transformer Heads & 8 (Discrete); 4 (Continuous) \\
 & Number of Backbone Transformer Layers (N) & 2 \\
 & Number of Policy/Value Transformer Heads & 4 \\
 & Number of Policy/Value Transformer Layers (N) & 1 \\
 & Dropout Rate (p) & 0.1 \\
 & Activation Function &  GELU \\
 & Reward/Value Bins & 101 (Continuous); 601 (Discrete) \\
\midrule

\multirow{10}{*}{Optimization}
 & Training Context Length (H) & 10 \\
 & Replay Ratio & 0.25 \\
 & Buffer Reanalyze Frequency & $0$ (Discrete); $1/100000$ (Continuous) \\
 & Batch Size & 64 \\
 & Optimizer & AdamW \\
 & Learning Rate & $1 \times 10^{-4}$ \\
 & Next Latent State Loss Coefficient & 10 \\
 & Reward Loss Coefficient & 1 (Discrete); 0.1 (Continuous) \\
 & Policy Loss Coefficient & 1 (Discrete); 0.1 (Continuous) \\
 & Value Loss Coefficient & 0.5 (Discrete); 0.1 (Continuous) \\
 & Policy Entropy Coefficient & $1 \times 10^{-4}$ \\
 & Weight Decay & $10^{-4}$ \\
 & Max Gradient Norm & 5 \\
 & Discount Factor & 0.997 (0.925 in Cube Pushing Task) \\
 & Soft Target Update Momentum & 0.05 \\
 & Hard Target Network Update Frequency & 100 \\
 & Temporal Difference (TD) Steps & 5 \\
\bottomrule
\end{tabular}
}
\end{table}

\section{Compute Resources}

In our setup, training COMET for 500k environment steps on a single NVIDIA H100 (80 GB) GPU takes approximately 18 hours on average across different tasks.

\section{COMET`s trajectory rollouts}
\label{app:rollout}
\begin{figure}[H]
\centering
\includegraphics[height=0.6\textheight]{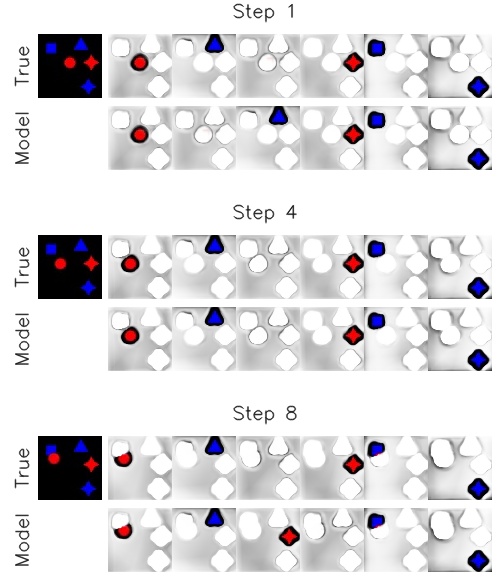}
\caption{Trajectory rollout generated using COMET’s policy for the Object Goal Task. The first row at each time step shows the real observation from the environment along with attention maps produced by the SLATE model over each slot inferred by the SLATE model. 
The second row shows attention maps produced by the SLATE model for each slot predicted by COMET’s dynamics model.  COMET’s dynamics model correctly predicted object representations.} 
\end{figure}

\begin{figure}[H] 
\centering
\includegraphics[height=0.6\textheight]{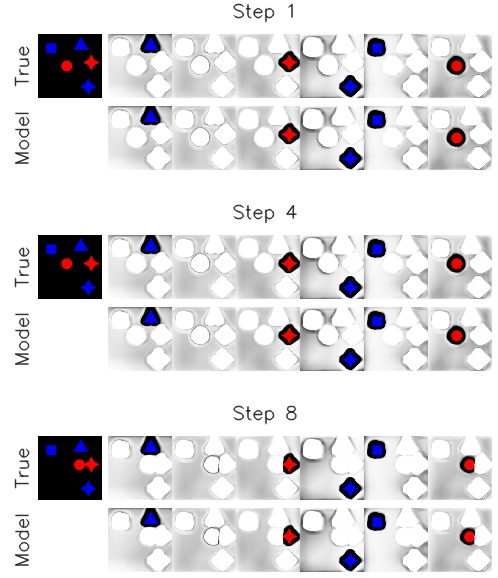}
\caption{Trajectory rollout generated using a random policy for the Object Goal Task. The first row at each time step shows the real observation from the environment along with attention maps produced by the SLATE model over each slot inferred by the SLATE model. 
The second row shows attention maps produced by the SLATE model for each slot predicted by COMET’s dynamics model. COMET’s dynamics model correctly predicted object representations.} 
\end{figure} 

\begin{figure}[H] 
\centering
\includegraphics[height=0.6\textheight]{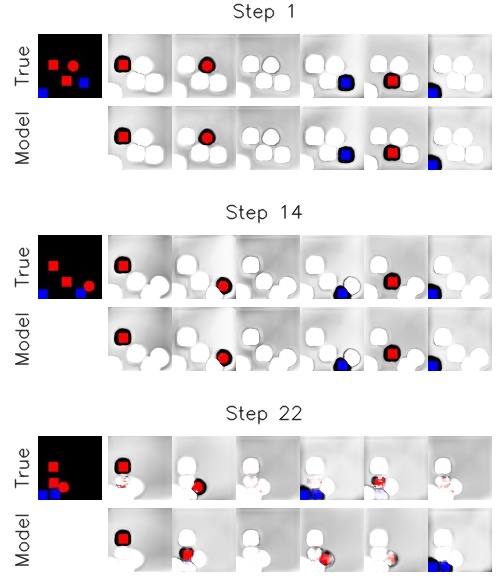}
\caption{Trajectory rollout generated using COMET’s policy for the Object Interaction Task. The first row at each time step shows the real observation from the environment along with attention maps produced by the SLATE model over each slot inferred by the SLATE model. The second row shows attention maps produced by the SLATE model for each slot predicted by COMET’s dynamics model. COMET’s dynamics model correctly predicted object representations; however, near the end of the trajectory, when objects became spatially close, the model produced inaccurate predictions. Notably, the ground-truth slots in this situation also became inconsistent with those from previous steps.} 
\end{figure}

\begin{figure}[H] 
\centering
\includegraphics[height=0.6\textheight]{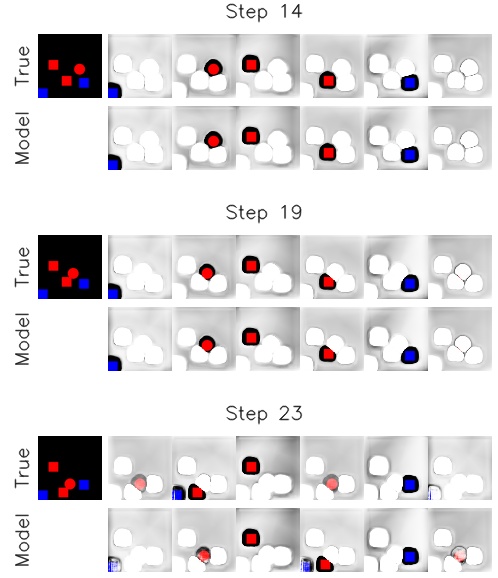}
\caption{Trajectory rollout generated using a random policy for the Object Interaction Task. The first row at each time step shows the real observation from the environment along with attention maps produced by the SLATE model over each slot inferred by the SLATE model. The second row shows attention maps produced by the SLATE model for each slot predicted by COMET’s dynamics model. COMET’s dynamics model correctly predicted object representations; however, near the end of the trajectory, when objects became spatially close, the model produced inaccurate predictions. Notably, the ground-truth slots in this situation also became inconsistent with those from previous steps.} 
\end{figure}

\begin{figure}[H] 
\centering
\includegraphics[height=0.6\textheight]{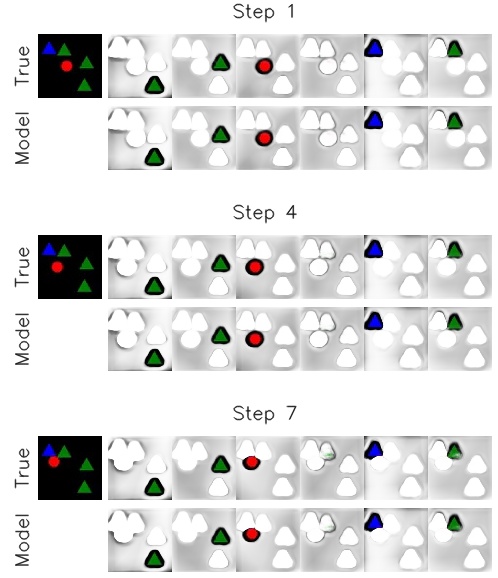}
\caption{Trajectory rollout generated using COMET’s policy for the Object Comparison Task. The first row at each time step shows the real observation from the environment along with attention maps produced by the SLATE model over each slot inferred by the SLATE model. The second row shows attention maps produced by the SLATE model for each slot predicted by COMET’s dynamics model. COMET’s dynamics model correctly predicted object representations.} 
\end{figure}

\begin{figure}[H] 
\centering
\includegraphics[height=0.6\textheight]{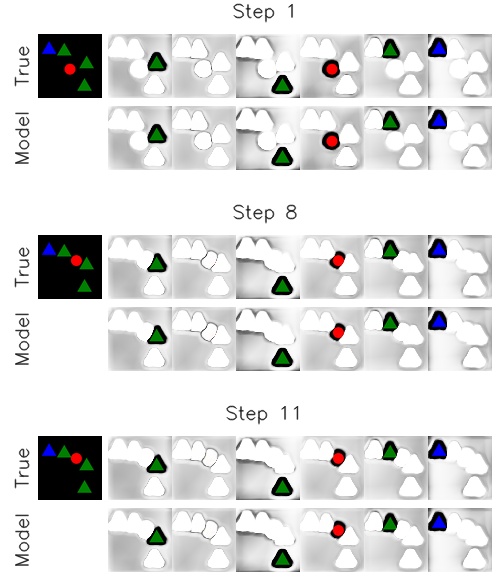}
\caption{Trajectory rollout generated using a random policy for the Object Comparison Task. The first row at each time step shows the real observation from the environment along with attention maps produced by the SLATE model over each slot inferred by the SLATE model. The second row shows attention maps produced by the SLATE model for each slot predicted by COMET’s dynamics model. COMET’s dynamics model correctly predicted object representations.} 
\end{figure} 

\begin{figure}[H] 
\centering
\includegraphics[height=0.6\textheight]{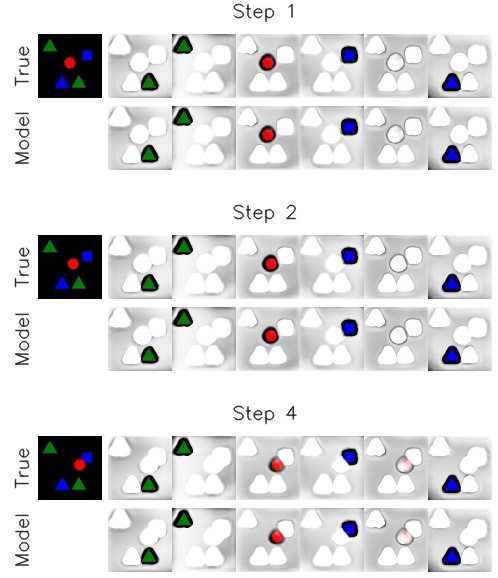}
\caption{Trajectory rollout generated using COMET’s policy for the Property Comparison Task. The first row at each time step shows the real observation from the environment along with attention maps produced by the SLATE model over each slot inferred by the SLATE model. The second row shows attention maps produced by the SLATE model for each slot predicted by COMET’s dynamics model. COMET’s dynamics model correctly predicted object representations.} 
\end{figure}

\begin{figure}[H] 
\centering
\includegraphics[height=0.6\textheight]{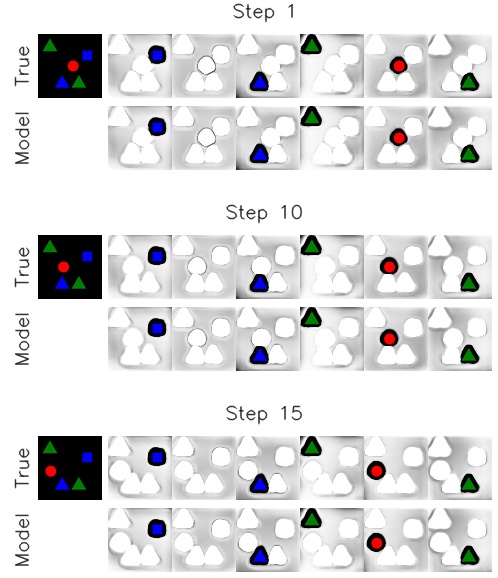}
\caption{Trajectory rollout generated using a random policy for the Property Comparison Task. The first row at each time step shows the real observation from the environment along with attention maps produced by the SLATE model over each slot inferred by the SLATE model. The second row shows attention maps produced by the SLATE model for each slot predicted by COMET’s dynamics model. COMET’s dynamics model correctly predicted object representations.} 
\end{figure} 

\begin{figure}[H] 
\includegraphics[width=\textwidth]{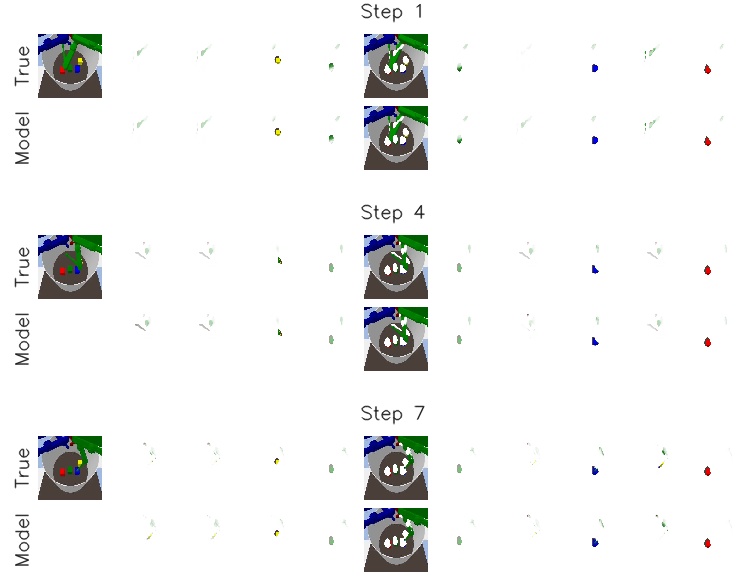}
\caption{Trajectory rollout generated using COMET’s policy for the Object Reaching Task. The first row at each time step shows the real observation from the environment along with attention maps produced by the SLATE model over each slot inferred by the SLATE model. The second row shows attention maps produced by the SLATE model for each slot predicted by COMET’s dynamics model. COMET’s dynamics model correctly predicted object representations.} 
\end{figure}

\begin{figure}[H] 
\includegraphics[width=\textwidth]{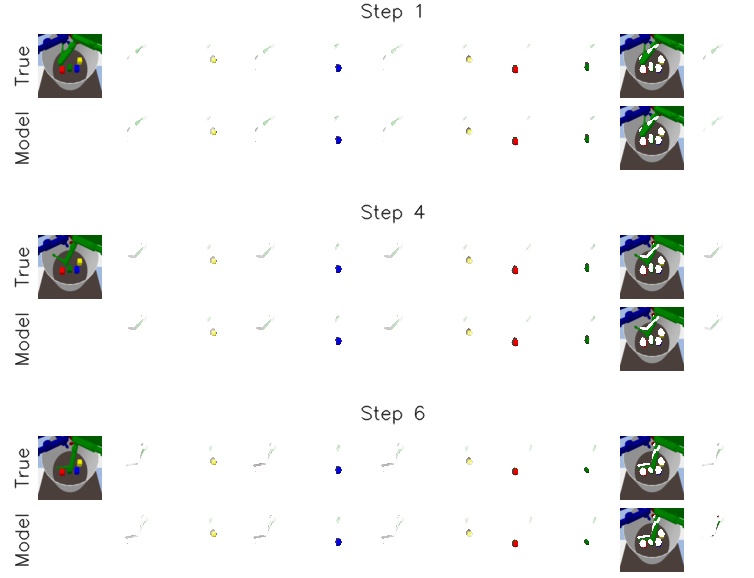}
\caption{Trajectory rollout generated using a random policy for the Object Reaching Task. The first row at each time step shows the real observation from the environment along with attention maps produced by the SLATE model over each slot inferred by the SLATE model. The second row shows attention maps produced by the SLATE model for each slot predicted by COMET’s dynamics model. COMET’s dynamics model correctly predicted object representations.} 
\end{figure}

\begin{figure}[H] 
\includegraphics[width=\textwidth]{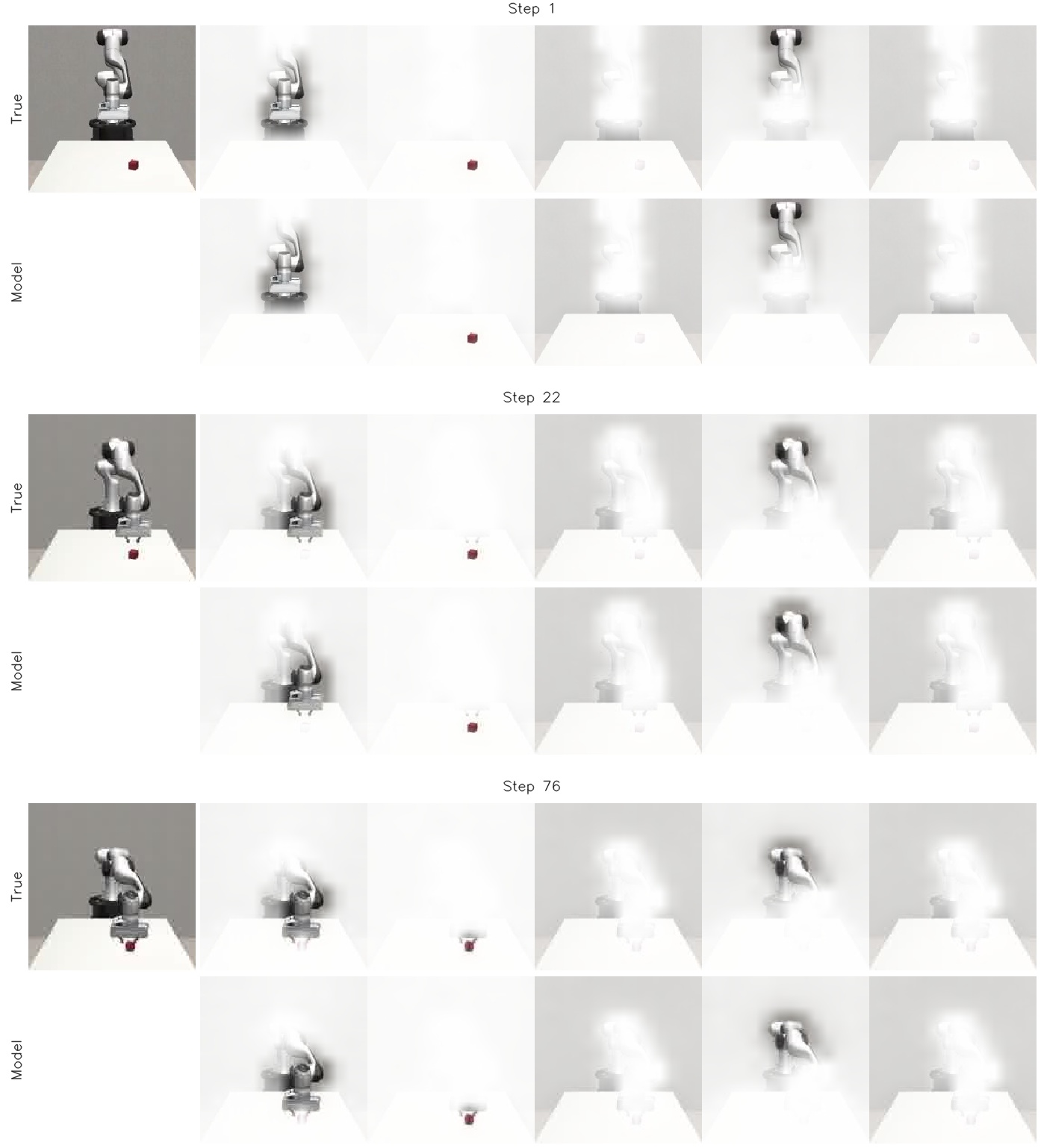}
\caption{Trajectory rollout generated using COMET’s policy for the Block Lifting Task. The first row at each time step shows the real observation from the environment along with attention maps produced by the DINOSAUR model over each slot inferred by the DINOSAUR model. The second row shows attention maps produced by the DINOSAUR model for each slot predicted by COMET’s dynamics mode. COMET’s dynamics model correctly predicted object representations. COMET’s dynamics model correctly predicted object representations.} 
\end{figure}

\begin{figure}[H] 
\includegraphics[width=\textwidth]{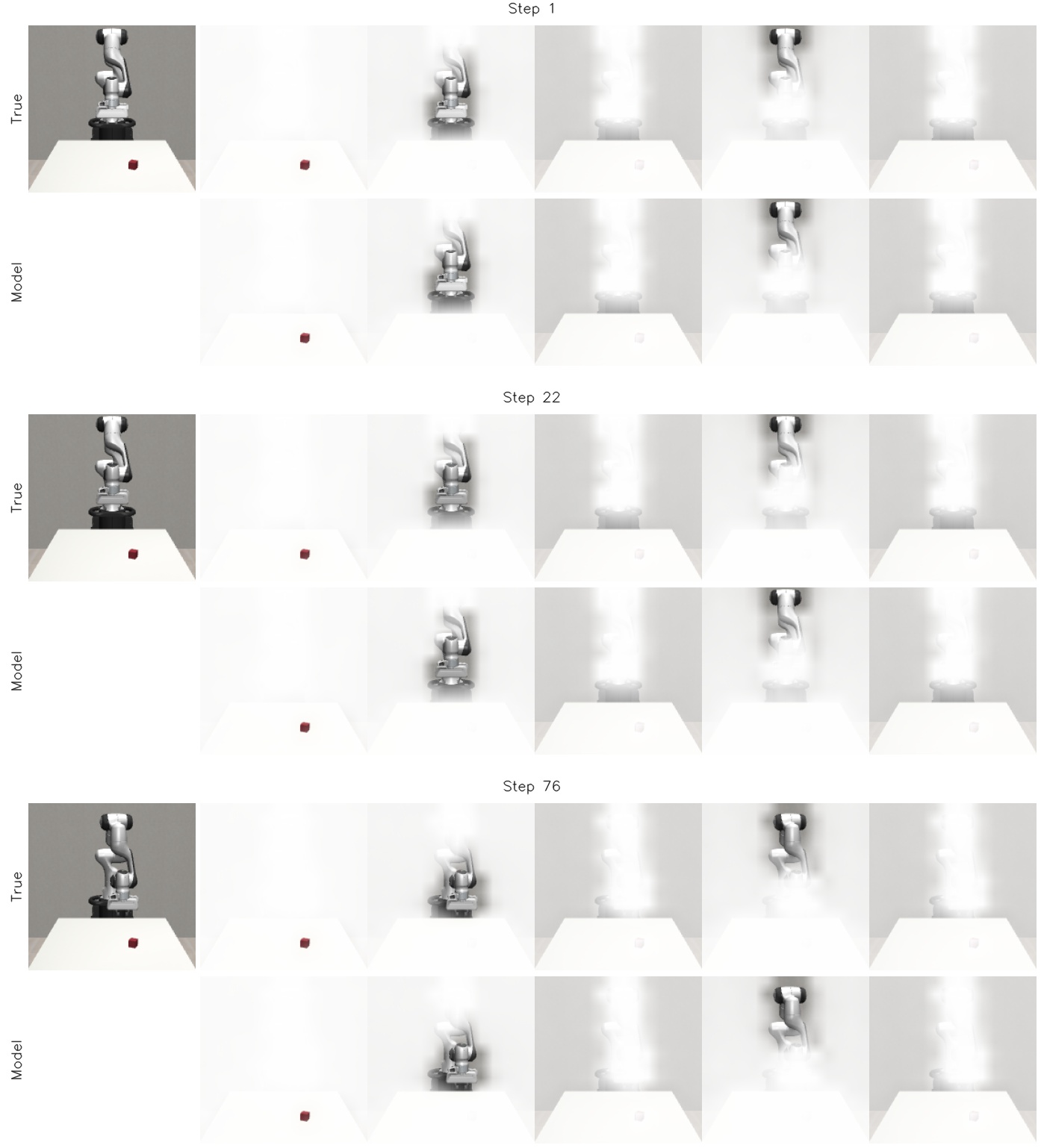}
\caption{Trajectory rollout generated using a random policy for the Block Lifting Task. The first row at each time step shows the real observation from the environment along with attention maps produced by the DINOSAUR model over each slot inferred by the DINOSAUR encoder. The second row shows attention maps produced by the DINOSAUR model for each slot predicted by COMET’s dynamics mode. COMET’s dynamics model correctly predicted object representations.} 
\end{figure}

\begin{figure}[H] 
\centering
\includegraphics[height=0.8\textheight]{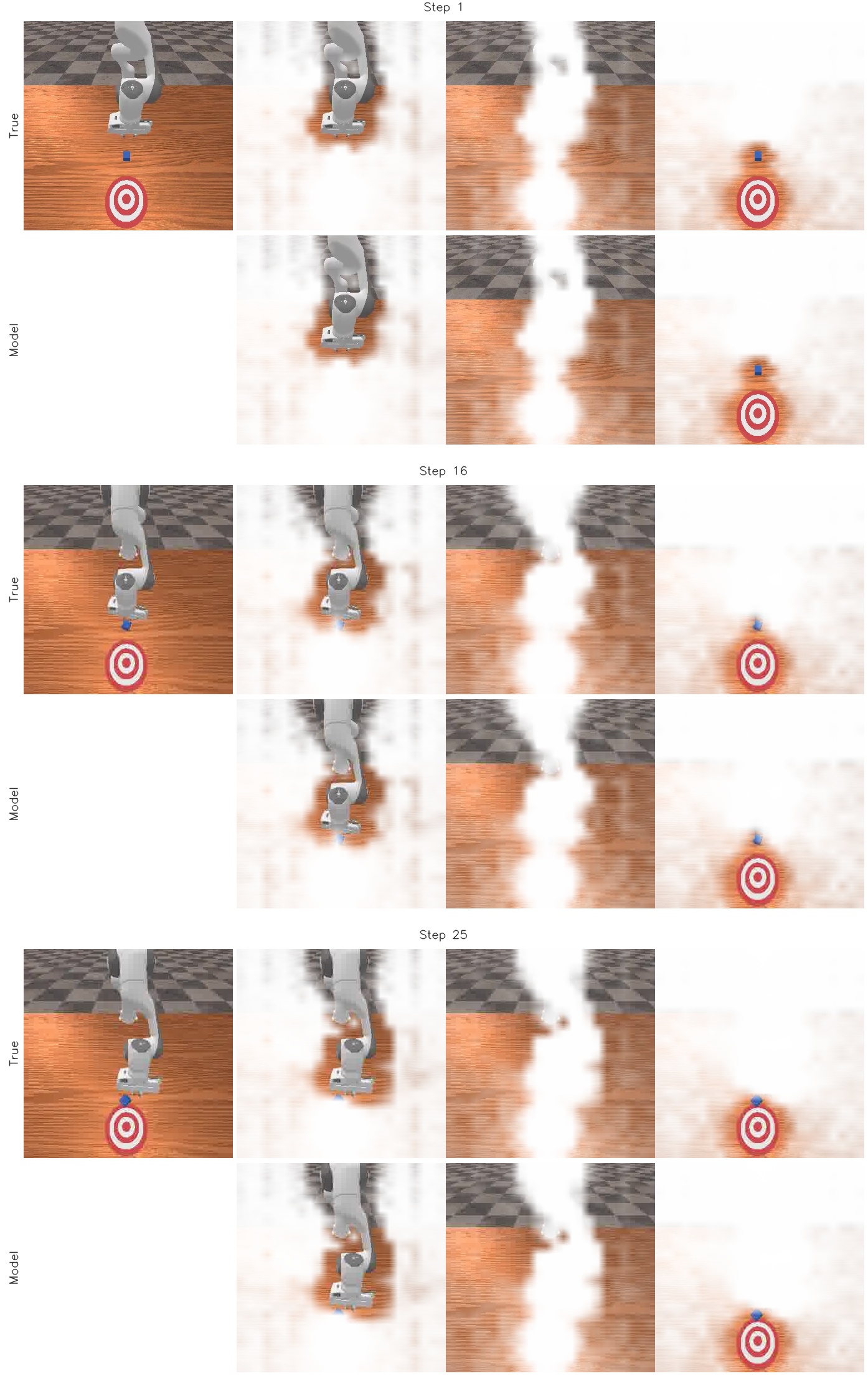}
\caption{Trajectory rollout generated using COMET’s policy for the Cube Pushing Task. The first row at each time step shows the real observation from the environment along with attention maps produced by the Slot Contrast model over each slot inferred by the Slot Contrast model. The second row shows attention maps produced by the Slot Contrast model for each slot predicted by COMET’s dynamics mode. COMET’s dynamics model correctly predicted object representations.} 
\end{figure}

\begin{figure}[H] 
\centering
\includegraphics[height=0.8\textheight]{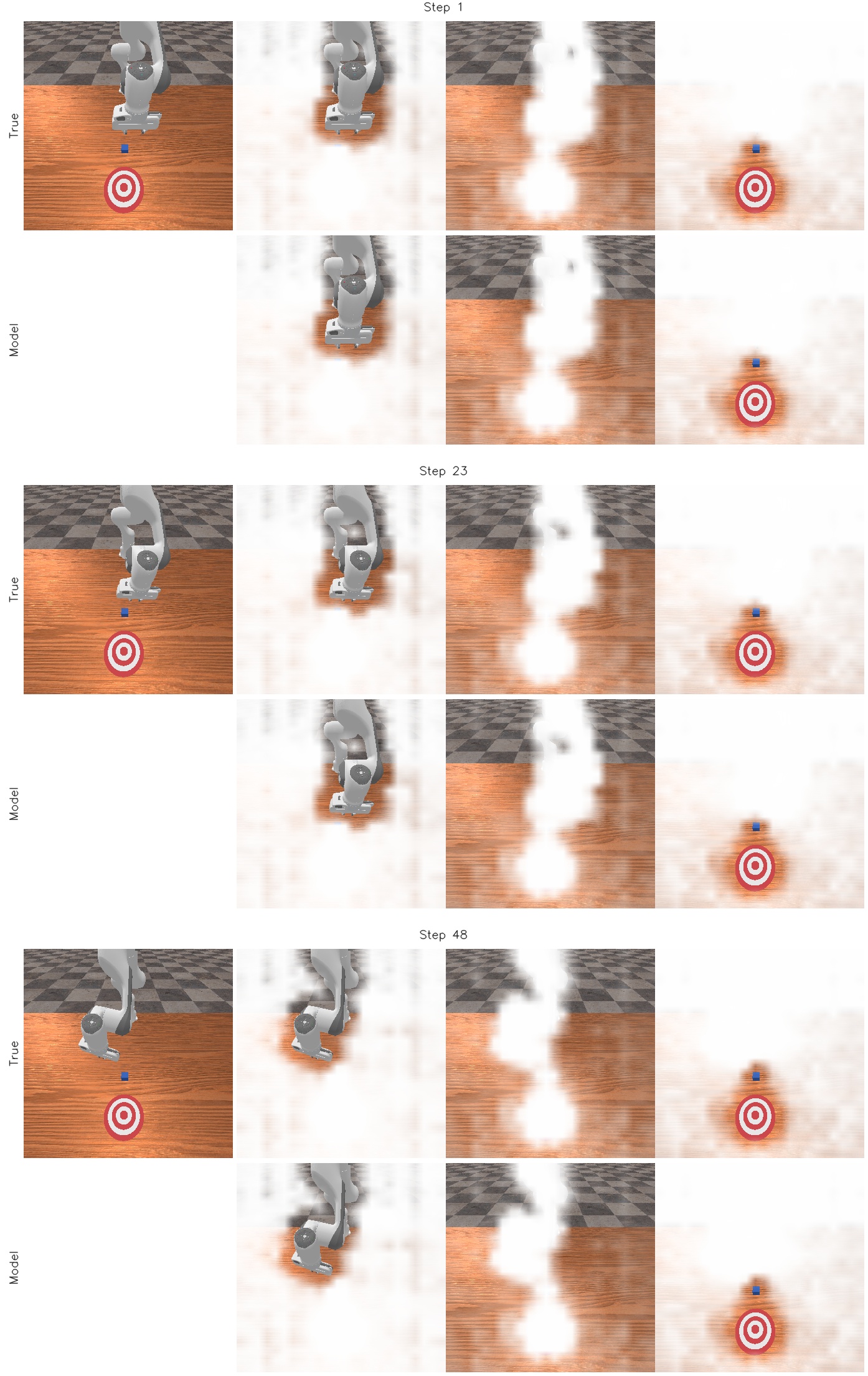}
\caption{Trajectory rollout generated using a random policy for the Cube Pushing Task. The first row at each time step shows the real observation from the environment along with attention maps produced by the Slot Contrast model over each slot inferred by the Slot Contrast model. The second row shows attention maps produced by the Slot Contrast model for each slot predicted by COMET’s dynamics mode. COMET’s dynamics model correctly predicted object representations.} 
\end{figure}

\begin{figure}[H] 
\centering
\includegraphics[width=0.6\textheight]{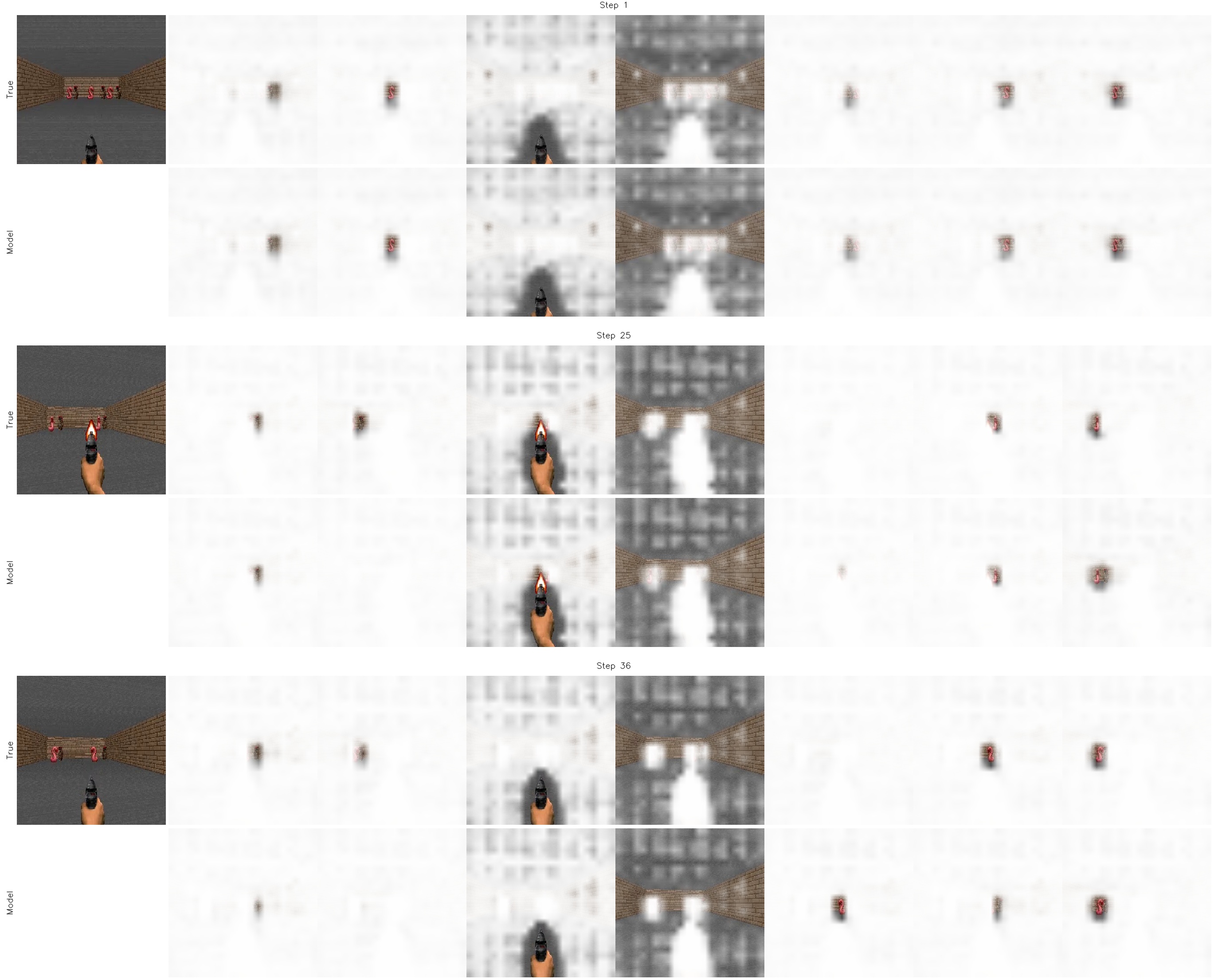}
\caption{Trajectory rollout generated using COMET’s policy for the Defend The Line Task. The first row at each time step shows the real observation from the environment along with attention maps produced by the Slot Contrast model over each slot inferred by the Slot Contrast model. The second row shows attention maps produced by the Slot Contrast model for each slot predicted by COMET’s dynamics model. COMET’s dynamics model correctly predicted object representations, with prediction errors appearing near the end of the trajectory.} 
\end{figure}

\begin{figure}[H] 
\centering
\includegraphics[width=0.6\textheight]{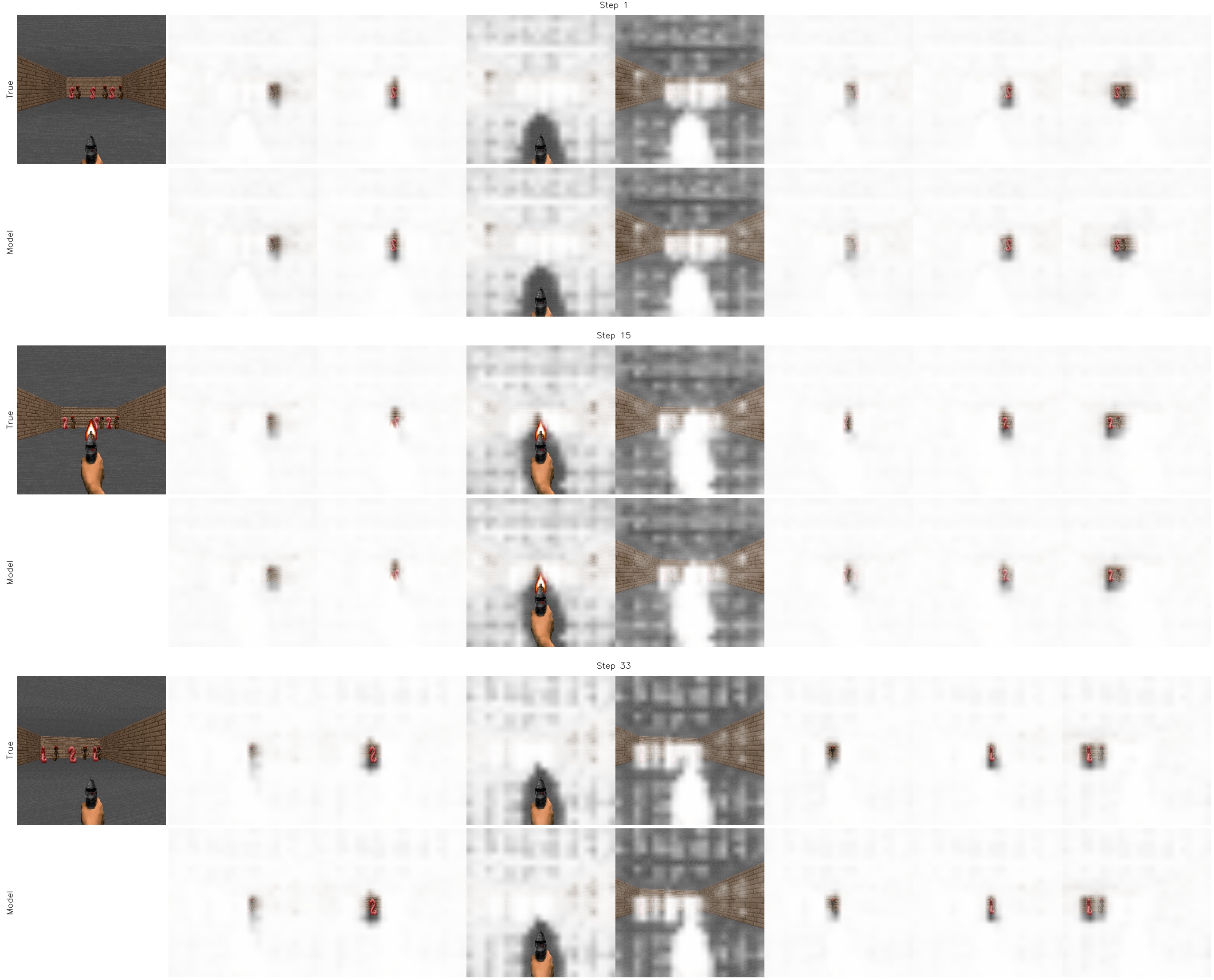}
\caption{Trajectory rollout generated using a random policy for the Defend The Line Task. The first row at each time step shows the real observation from the environment along with attention maps produced by the Slot Contrast model over each slot inferred by the Slot Contrast model. The second row shows attention maps produced by the Slot Contrast model for each slot predicted by COMET’s dynamics mode. COMET’s dynamics model correctly predicted object representations.} 
\end{figure} 

\section{COMET`s causality probabilities in policy and value models}
\label{app:probs}
\begin{figure}[H] 
\includegraphics[width=\textwidth]{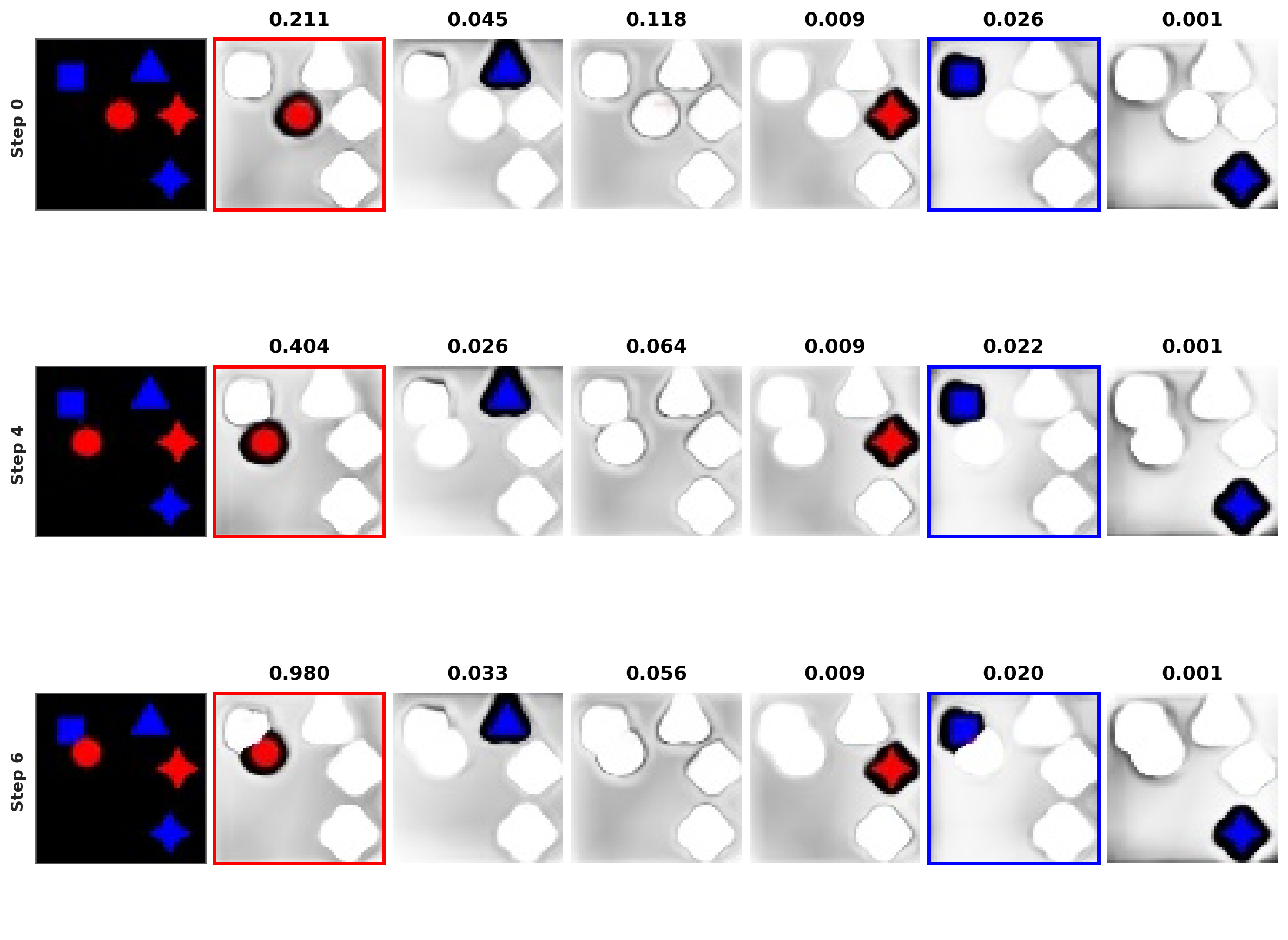}
\caption{
Per-slot causality scores for the value transformer in the Object Goal Task. Each row corresponds to a time step and shows the real observation from the environment together with attention maps produced by the SLATE model for each slot inferred by the SLATE model. The number above each slot denotes its causality score $\alpha_t^i$, indicating the probability that the corresponding object is causally relevant for value prediction. Red bounding boxes indicate agent-related objects, blue bounding boxes indicate target object. The causality score is highest for the agent object across all time steps.
} 
\end{figure}

\begin{figure}[H] 
\includegraphics[width=\textwidth]{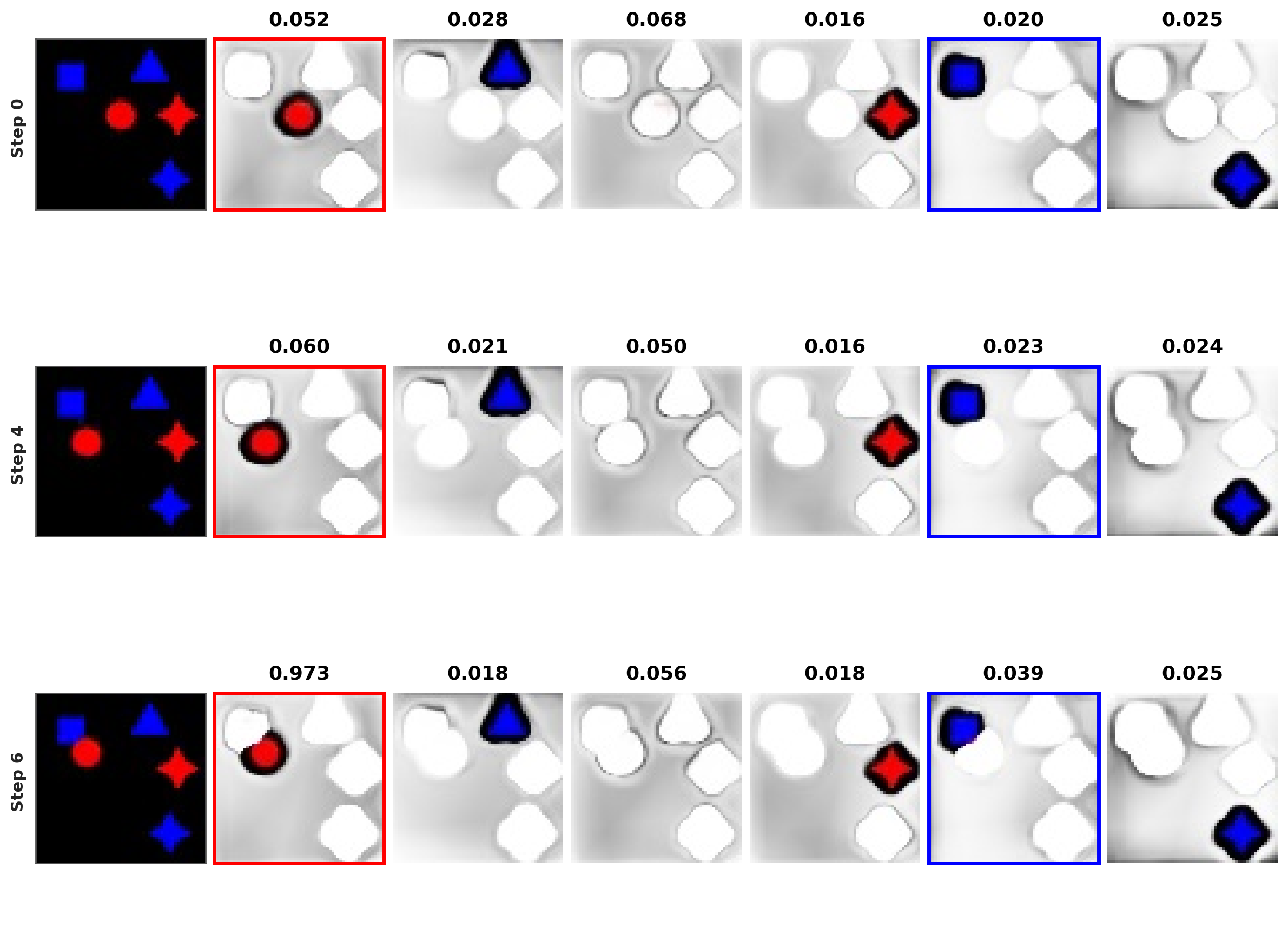}
\caption{Per-slot causality scores for the policy transformer in the Object Goal Task. Each row corresponds to a time step and shows the real observation from the environment together with attention maps produced by the SLATE model for each slot inferred by the SLATE model. The number above each slot denotes its causality score $\alpha_t^i$, indicating the probability that the corresponding object is causally relevant for policy prediction. Red bounding boxes indicate agent-related objects, blue bounding boxes indicate target object. The causality score is highest for the agent object across most time steps.} 
\end{figure} 

\begin{figure}[H] 
\includegraphics[width=\textwidth]{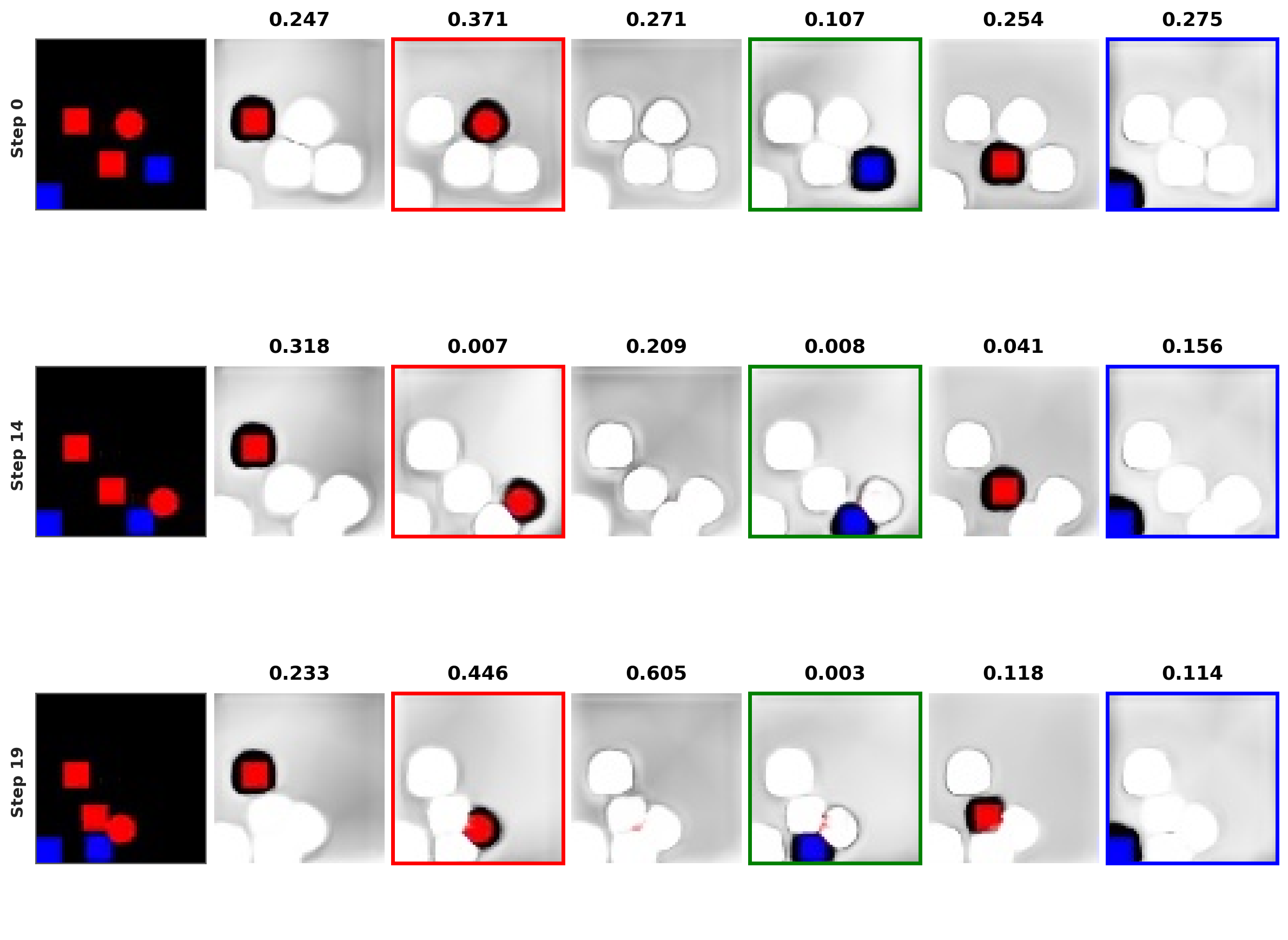}
\caption{
Per-slot causality scores for the value transformer in the Object interaction Task. Each row corresponds to a time step and shows the real observation from the environment together with attention maps produced by the SLATE model for each slot inferred by the SLATE model. The number above each slot denotes its causality score $\alpha_t^i$, indicating the probability that the corresponding object is causally relevant for value prediction. Red bounding boxes indicate agent-related objects, blue bounding boxes indicate target object, and green bounding boxes indicate auxiliary object. The agent and target objects generally receive high causality scores; however, near the end of the trajectory, the model incorrectly assigns a higher causality score to a background object and produces comparable causality scores for other irrelevant objects.
} 
\end{figure}

\begin{figure}[H] 
\includegraphics[width=\textwidth]{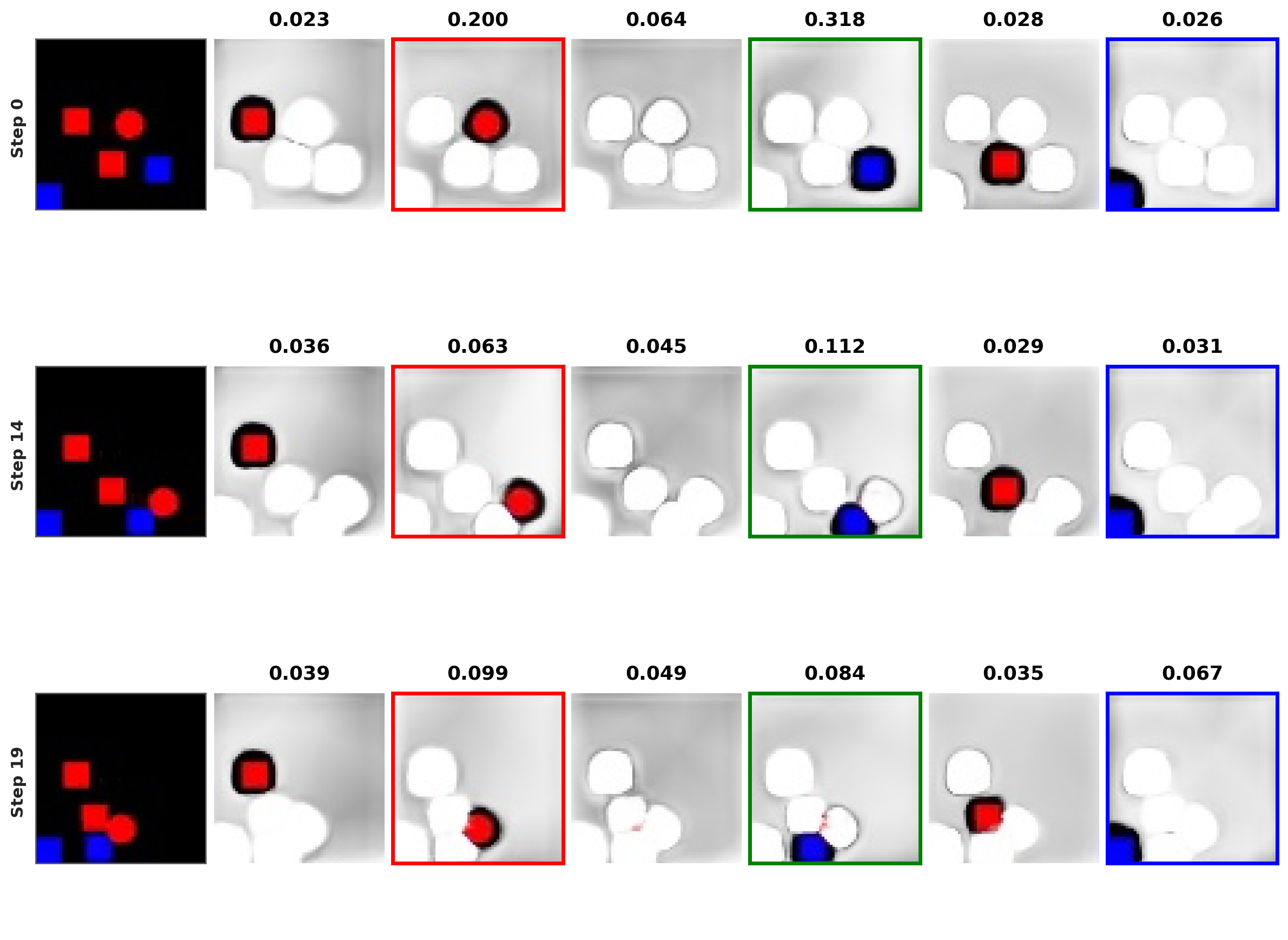}
\caption{Per-slot causality scores for the policy transformer in the Object Interaction Task. Each row corresponds to a time step and shows the real observation from the environment together with attention maps produced by the SLATE model for each slot inferred by the SLATE model. The number above each slot denotes its causality score $\alpha_t^i$, indicating the probability that the corresponding object is causally relevant for policy prediction. Red bounding boxes indicate agent-related objects, blue bounding boxes indicate target object, and green bounding boxes indicate auxiliary object.}  
\end{figure}

\begin{figure}[H] 
\includegraphics[width=\textwidth]{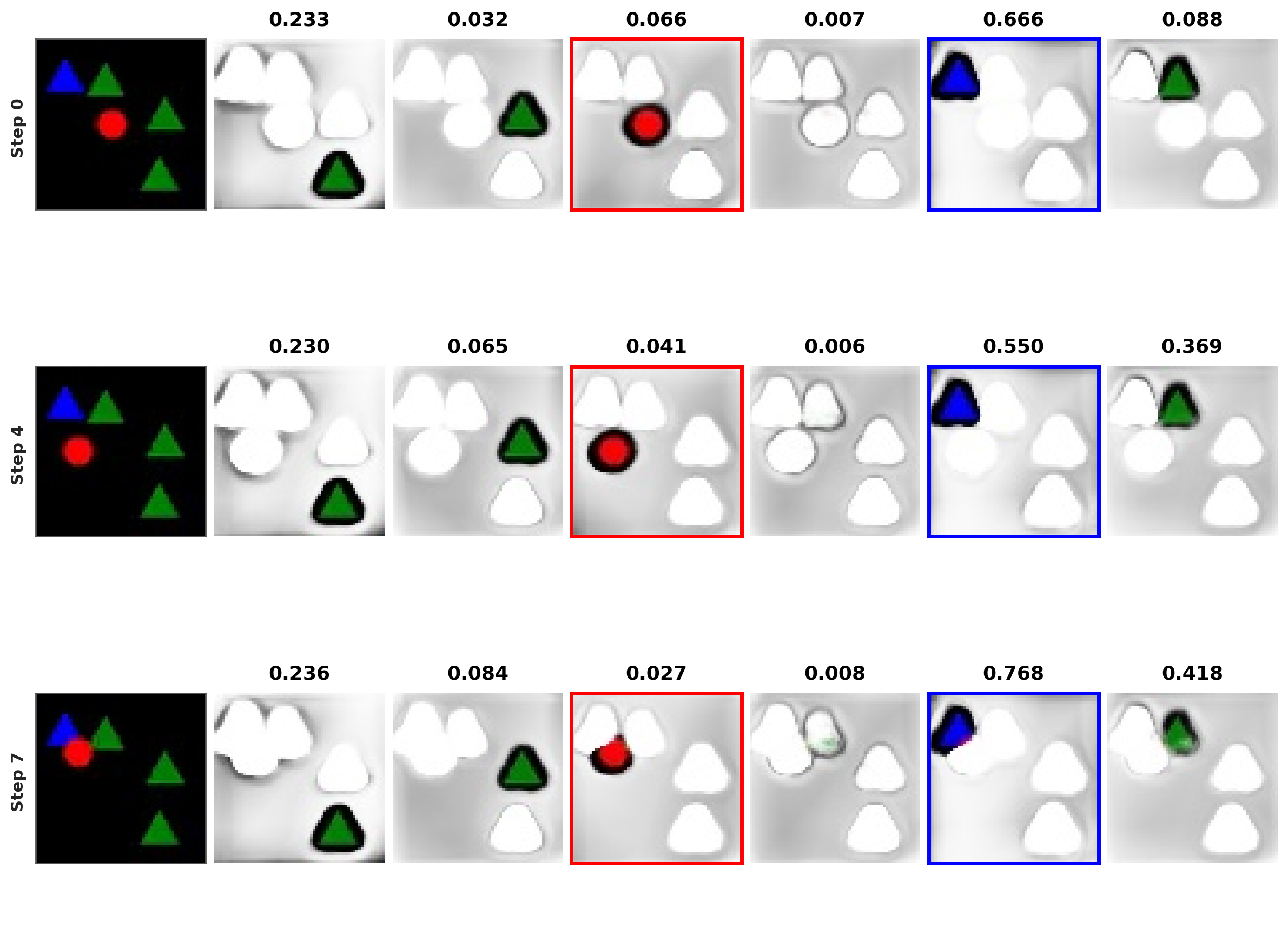}
\caption{
Per-slot causality scores for the value transformer in the Object Comparison Task. Each row corresponds to a time step and shows the real observation from the environment together with attention maps produced by the SLATE model for each slot inferred by the SLATE model. The number above each slot denotes its causality score $\alpha_t^i$, indicating the probability that the corresponding object is causally relevant for value prediction. Red bounding boxes indicate agent-related objects, blue bounding boxes indicate target object. The agent, auxiliary, and target objects receive higher causality scores compared to other objects across time steps. The causality score is highest for the target object across most time steps. Some other objects receive higher causality scores than the agent object.
} 
\end{figure}

\begin{figure}[H] 
\includegraphics[width=\textwidth]{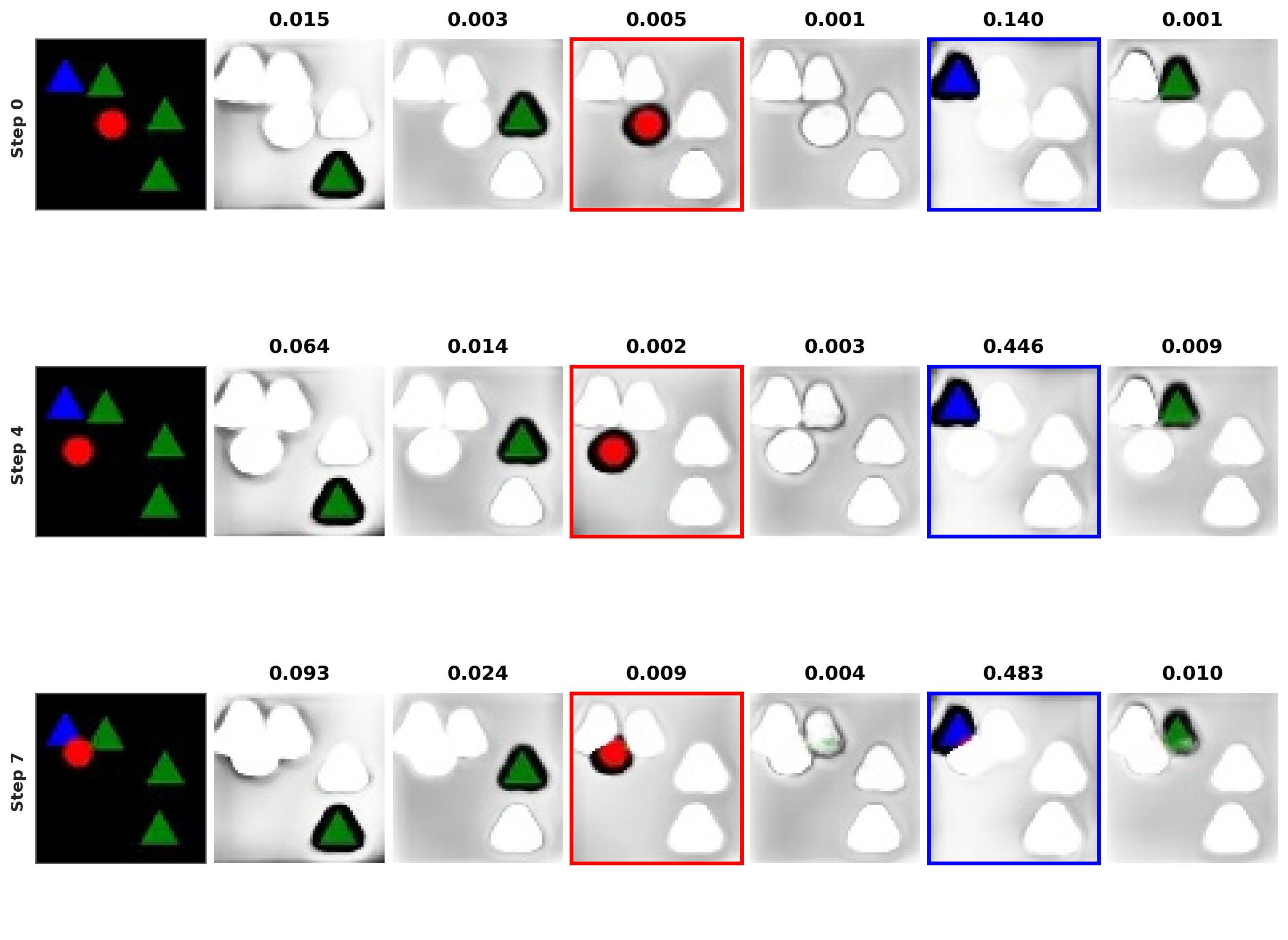}
\caption{Per-slot causality scores for the policy transformer in the Object Comparison Task. Each row corresponds to a time step and shows the real observation from the environment together with attention maps produced by the SLATE model for each slot inferred by the SLATE model. The number above each slot denotes its causality score $\alpha_t^i$, indicating the probability that the corresponding object is causally relevant for policy prediction. Red bounding boxes indicate agent-related objects, blue bounding boxes indicate target object. The causality score is highest for the target object across most time steps. Some other objects receive higher causality scores than the agent object.}  
\end{figure} 

\begin{figure}[H] 
\includegraphics[width=\textwidth]{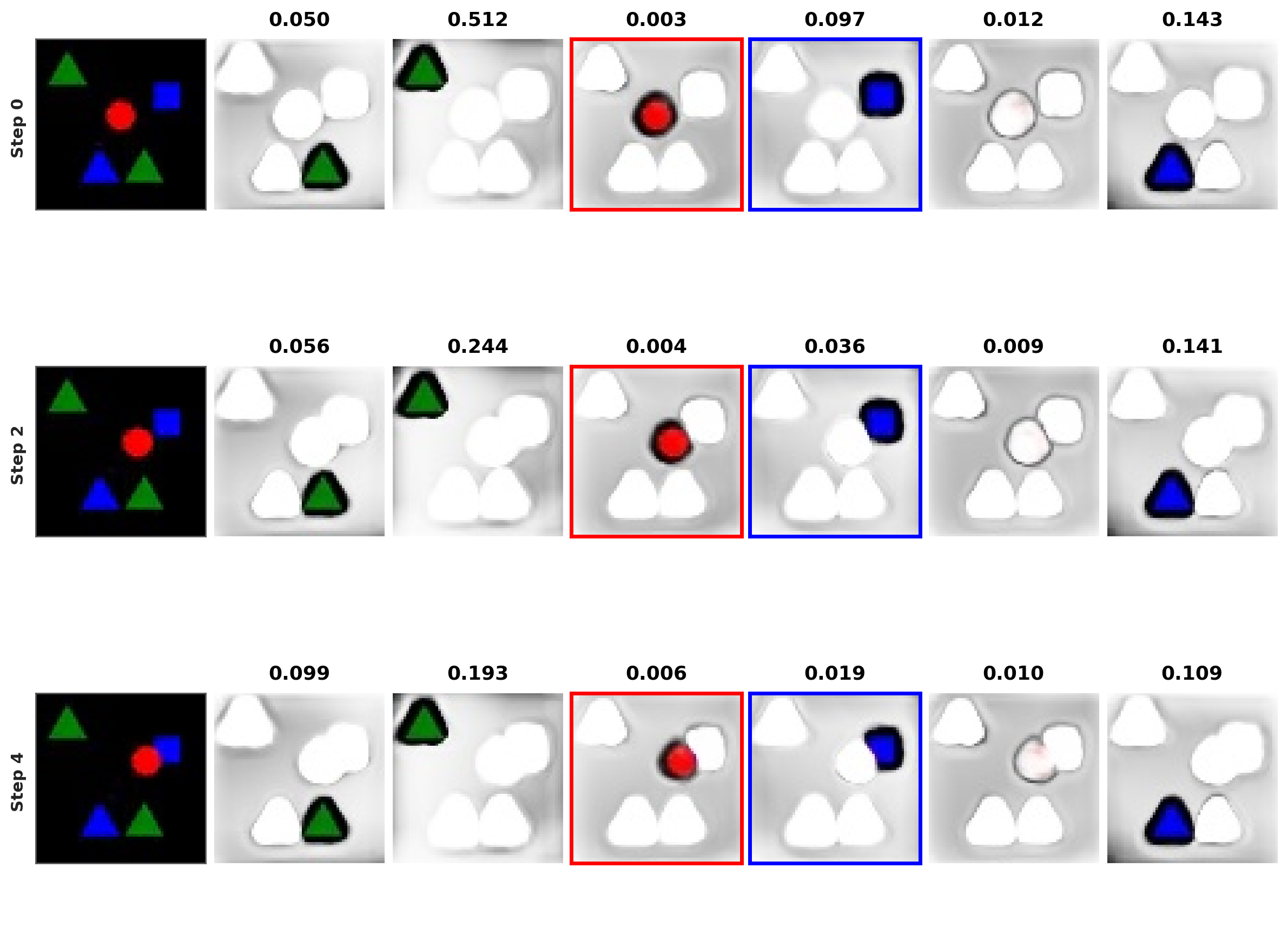}
\caption{
Per-slot causality scores for the value transformer in the Property Comparison Task. Each row corresponds to a time step and shows the real observation from the environment together with attention maps produced by the SLATE model for each slot inferred by the SLATE model. The number above each slot denotes its causality score $\alpha_t^i$, indicating the probability that the corresponding object is causally relevant for value prediction. Red bounding boxes indicate agent-related objects, blue bounding boxes indicate target object. In this case, the model incorrectly assigns causality scores to the agent and target objects relative to other objects.
} 
\end{figure}

\begin{figure}[H] 
\includegraphics[width=\textwidth]{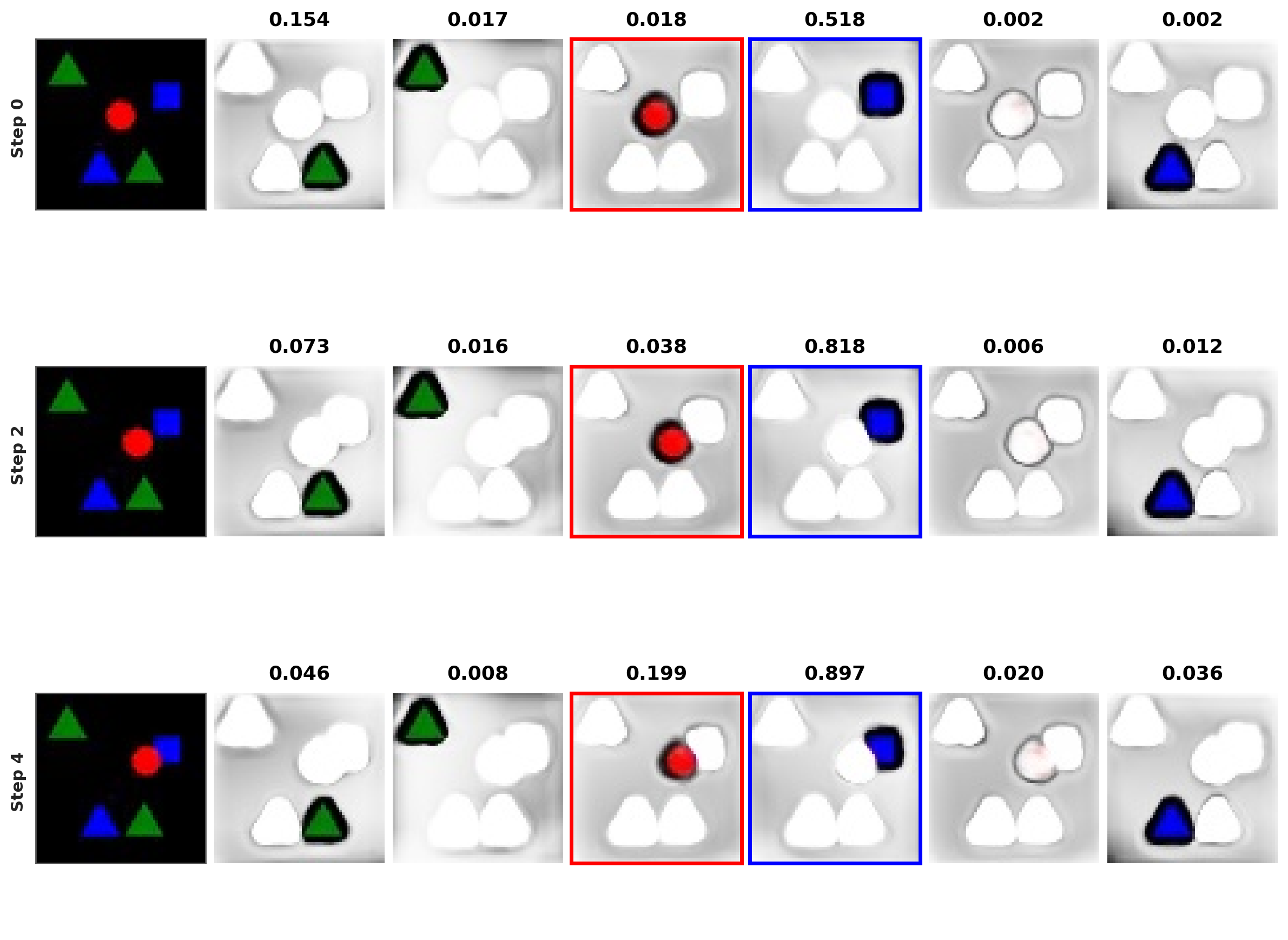}
\caption{Per-slot causality scores for the policy transformer in the Property Comparison Task. Each row corresponds to a time step and shows the real observation from the environment together with attention maps produced by the SLATE model for each slot inferred by the SLATE model. The number above each slot denotes its causality score $\alpha_t^i$, indicating the probability that the corresponding object is causally relevant for policy prediction. Red bounding boxes indicate agent-related objects, blue bounding boxes indicate target object. The causality score is highest for the target object across most time steps.}
\end{figure} 

\begin{figure}[H] 
\includegraphics[width=\textwidth]{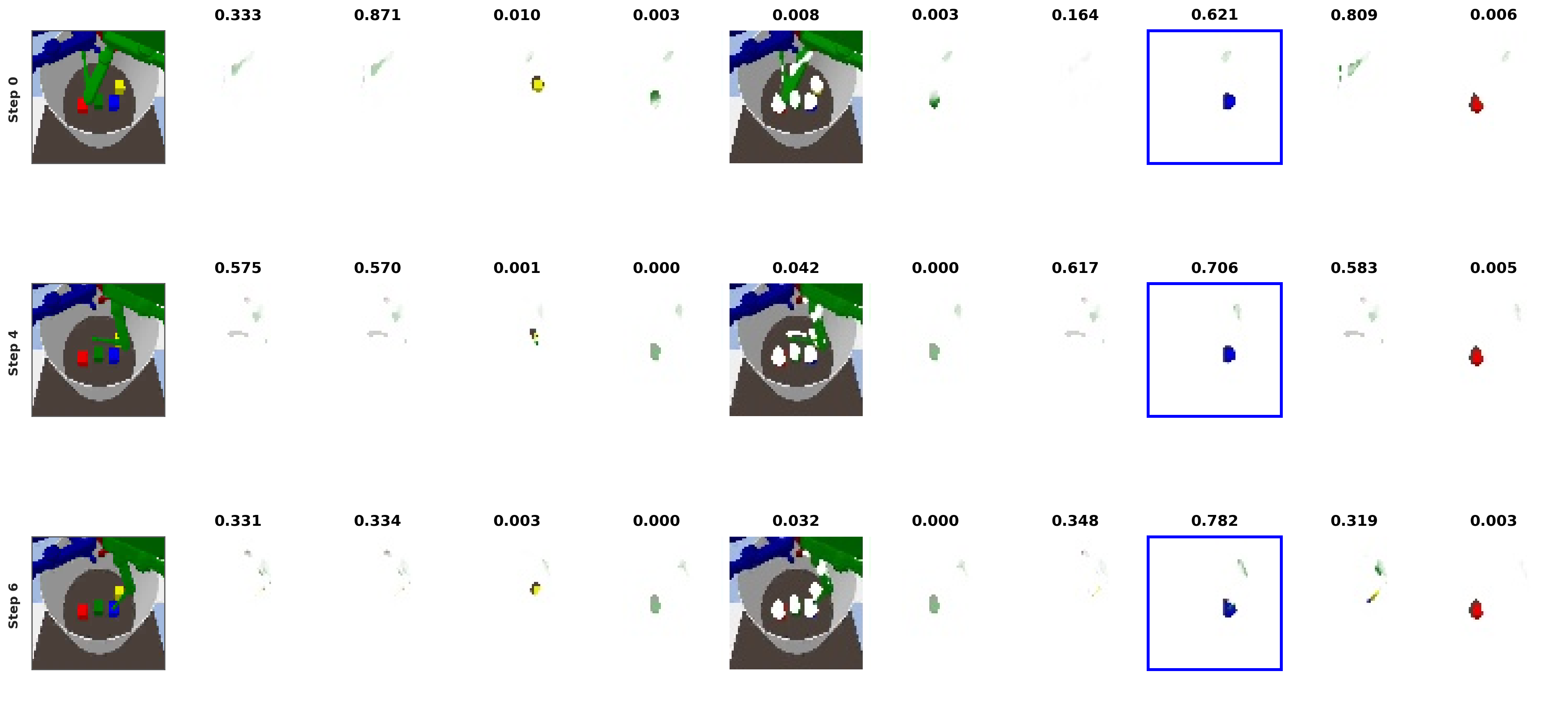}
\caption{
Per-slot causality scores for the value transformer in the Object Reaching Task. Each row corresponds to a time step and shows the real observation from the environment together with attention maps produced by the SLATE model for each slot inferred by the SLATE model. The number above each slot denotes its causality score $\alpha_t^i$, indicating the probability that the corresponding object is causally relevant for value prediction. Blue bounding boxes indicate target object.
} 
\end{figure}

\begin{figure}[H] 
\includegraphics[width=\textwidth]{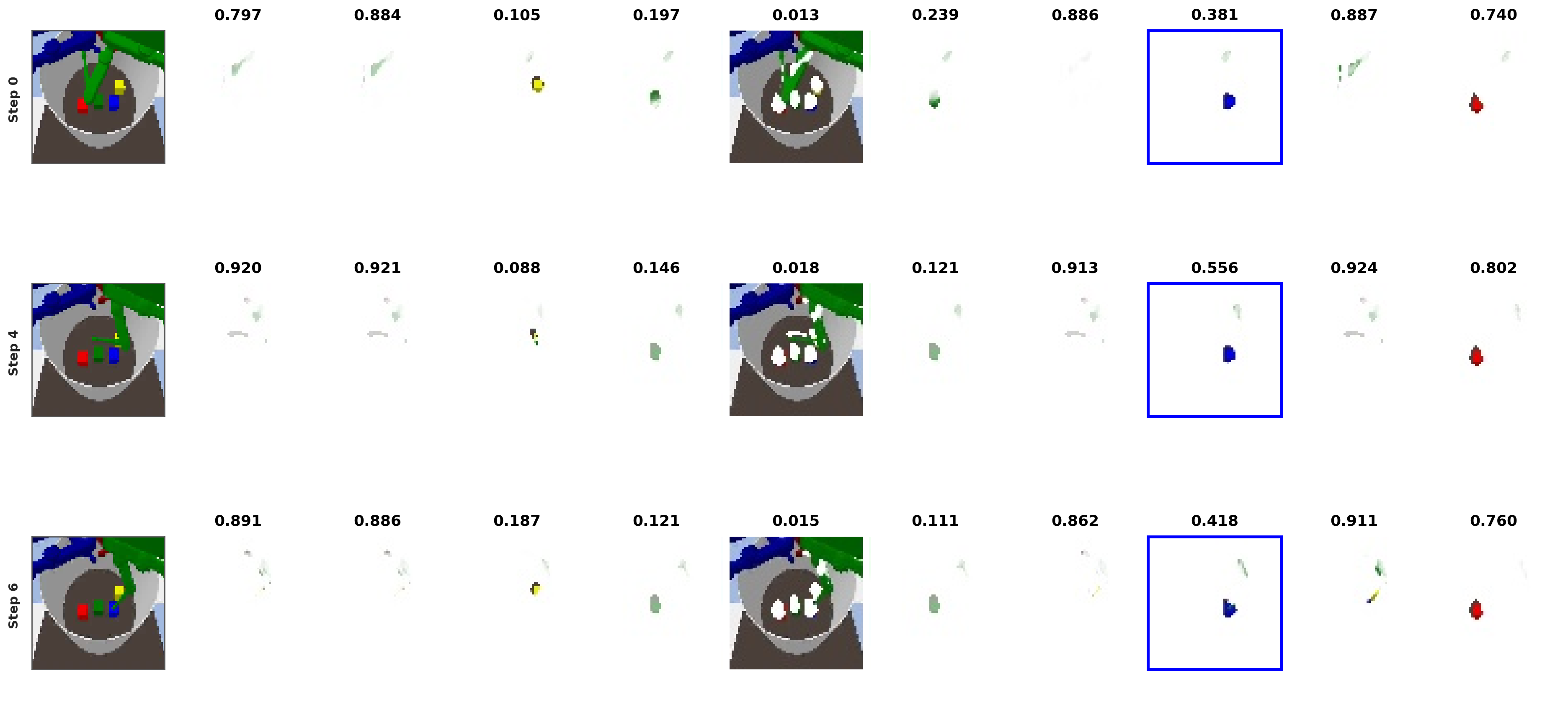}
\caption{Per-slot causality scores for the policy transformer in the Object Reaching Task. Each row corresponds to a time step and shows the real observation from the environment together with attention maps produced by the SLATE model for each slot inferred by the SLATE model. The number above each slot denotes its causality score $\alpha_t^i$, indicating the probability that the corresponding object is causally relevant for policy prediction. Red bounding boxes indicate agent-related objects, blue bounding boxes indicate target object. The causality score is highest for the target object and and for agent`s objects distributed across multiple slots. }
\end{figure}

\begin{figure}[H] 
\includegraphics[width=\textwidth]{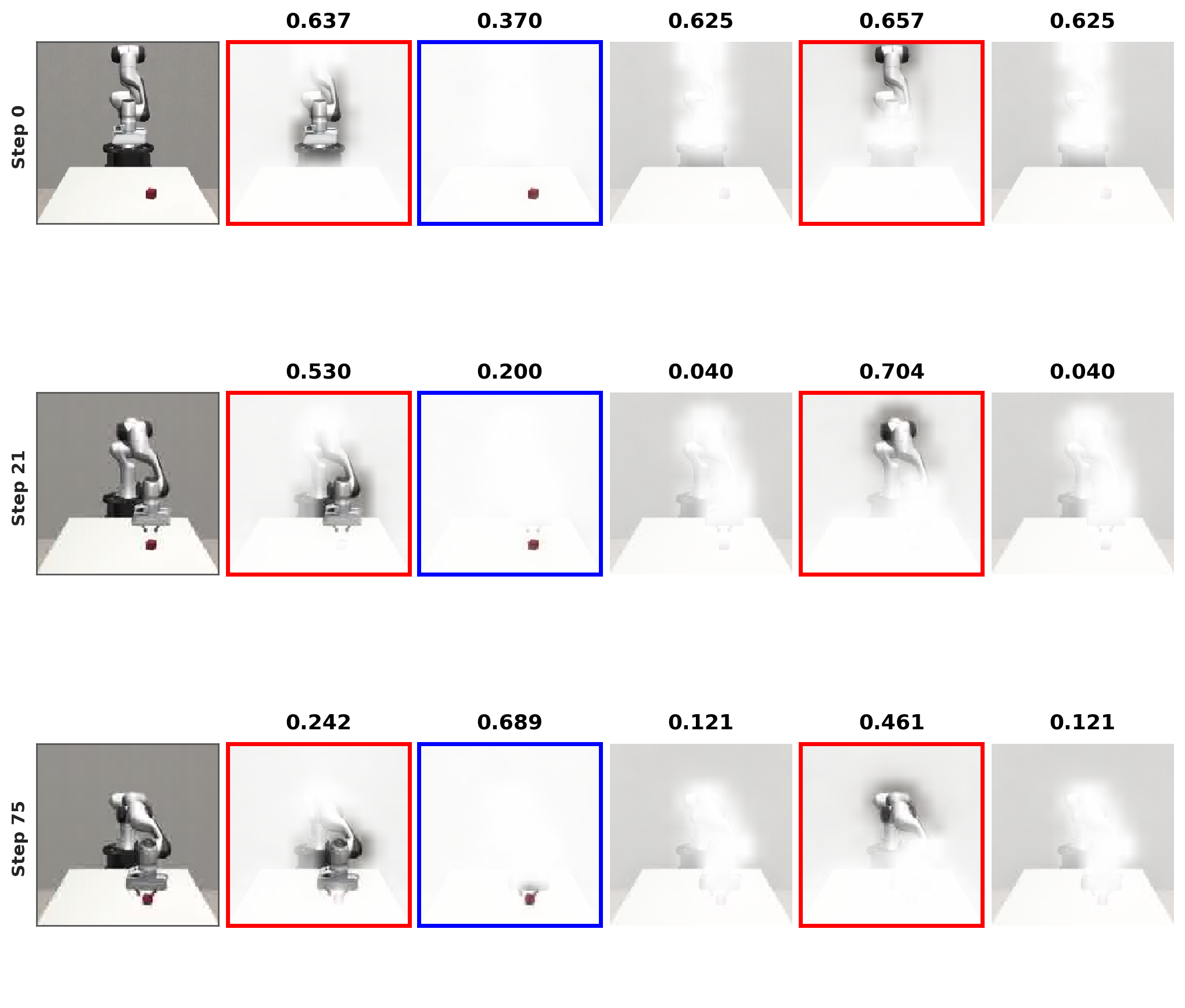}
\caption{
Per-slot causality scores for the value transformer in the Block Lifting Task. Each row corresponds to a time step and shows the real observation from the environment together with attention maps produced by the DINOSAUR model for each slot inferred by the DINOSAUR model. The number above each slot denotes its causality score $\alpha_t^i$, indicating the probability that the corresponding object is causally relevant for value prediction. Blue bounding boxes indicate target object. The causality score is highest for the target object and and for agent`s objects distributed across multiple slots. The causality score is highest for the target and agent objects across most time steps.
}  
\end{figure}

\begin{figure}[H] 
\includegraphics[width=\textwidth]{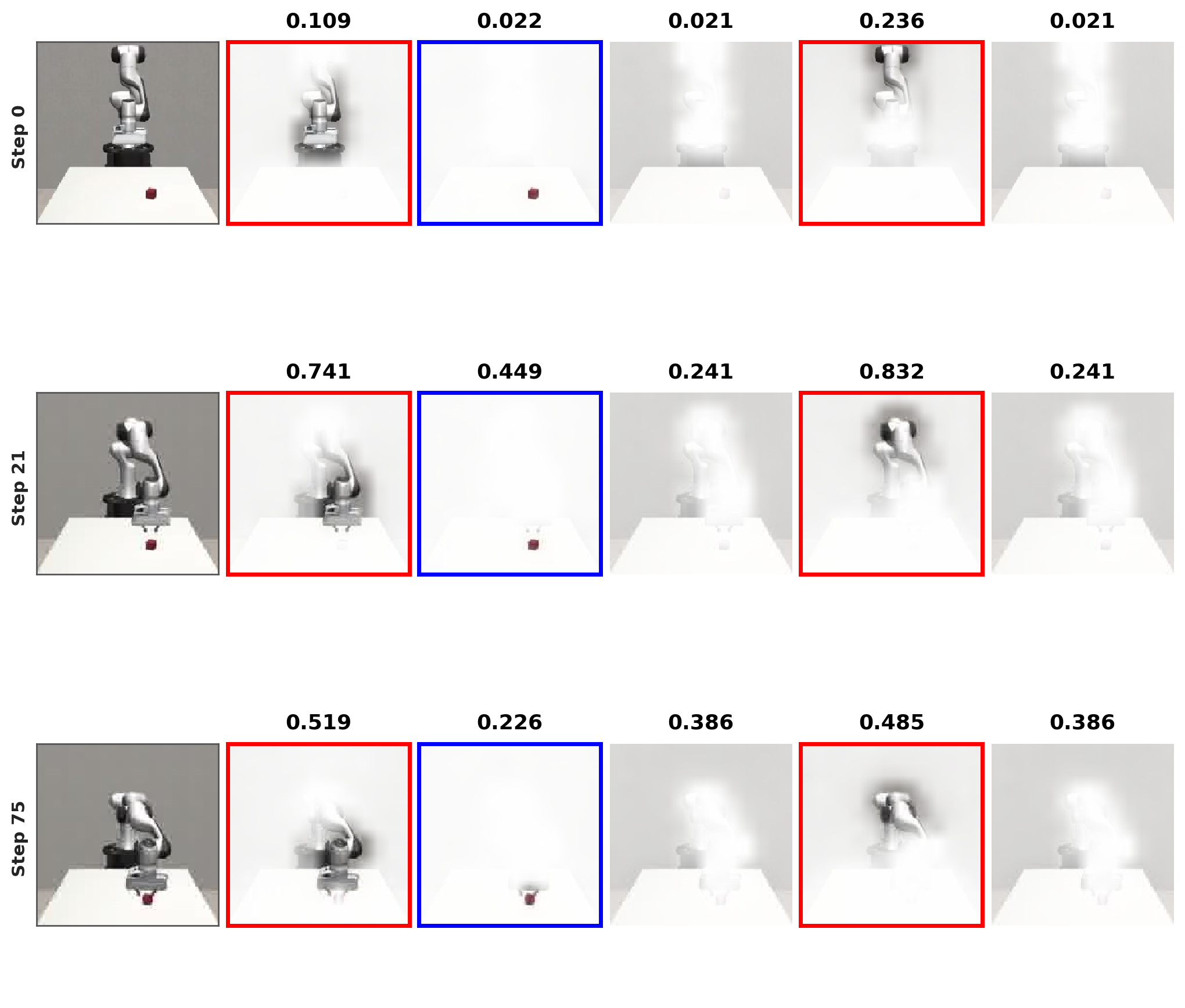}
\caption{Per-slot causality scores for the policy transformer in the Block Lifting Task. Each row corresponds to a time step and shows the real observation from the environment together with attention maps produced by the DINOSAUR model for each slot inferred by the DINOSAUR model. The number above each slot denotes its causality score $\alpha_t^i$, indicating the probability that the corresponding object is causally relevant for policy prediction. Red bounding boxes indicate agent-related objects, blue bounding boxes indicate target objects. The causality score is highest for the target and agent objects across most time steps.}
\end{figure}

\begin{figure}[H] 
\includegraphics[width=\textwidth]{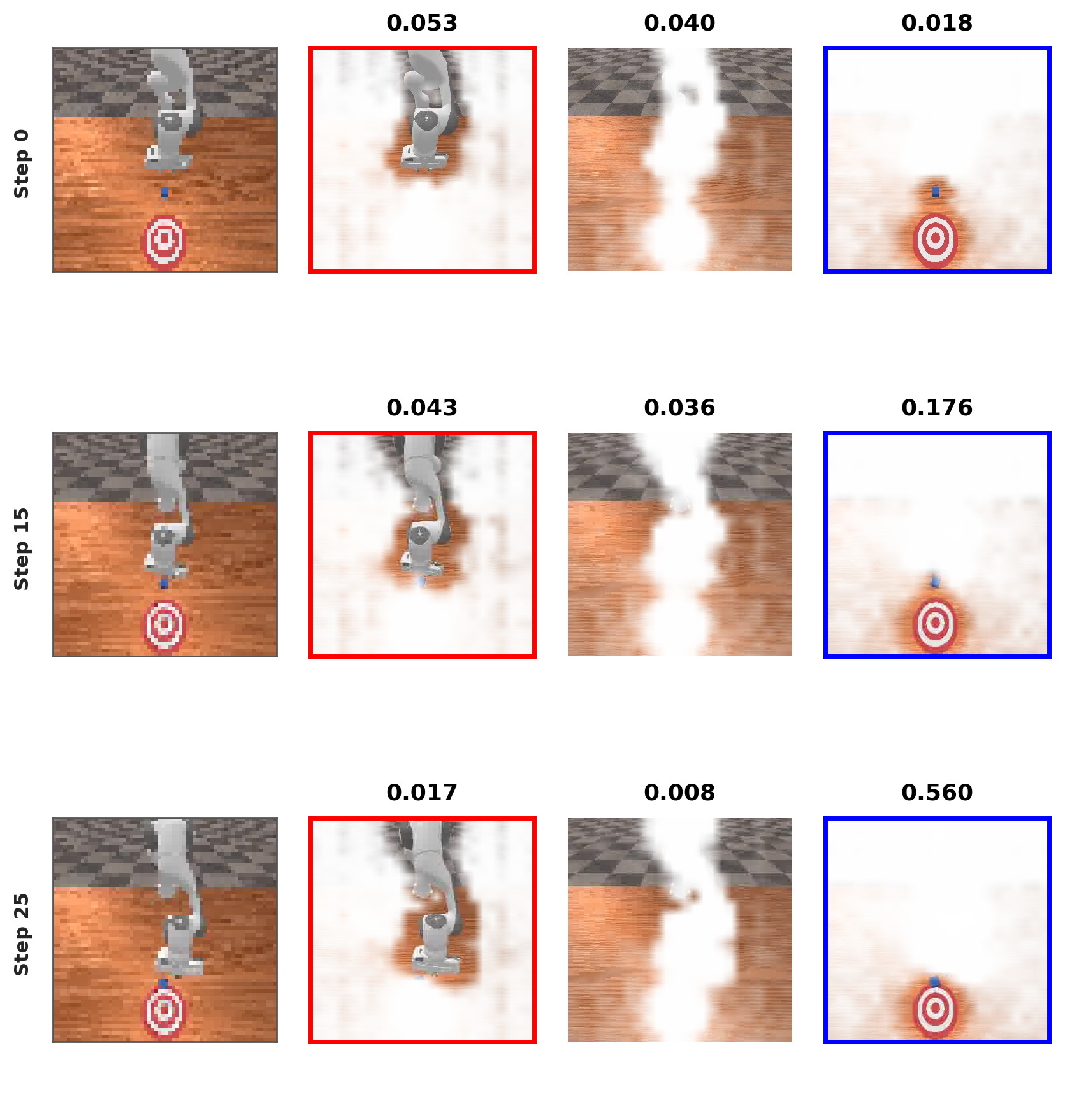}
\caption{
Per-slot causality scores for the value transformer in the Cube Pushing Task. Each row corresponds to a time step and shows the real observation from the environment together with attention maps produced by the Slot Contrast model for each slot inferred by the Slot Contrast model. The number above each slot denotes its causality score $\alpha_t^i$, indicating the probability that the corresponding object is causally relevant for value prediction. Red bounding boxes indicate agent-related objects,
blue bounding boxes indicate target objects. The causality score is highest for the target and agent objects across most time steps.
}
\end{figure}

\begin{figure}[H] 
\includegraphics[width=\textwidth]{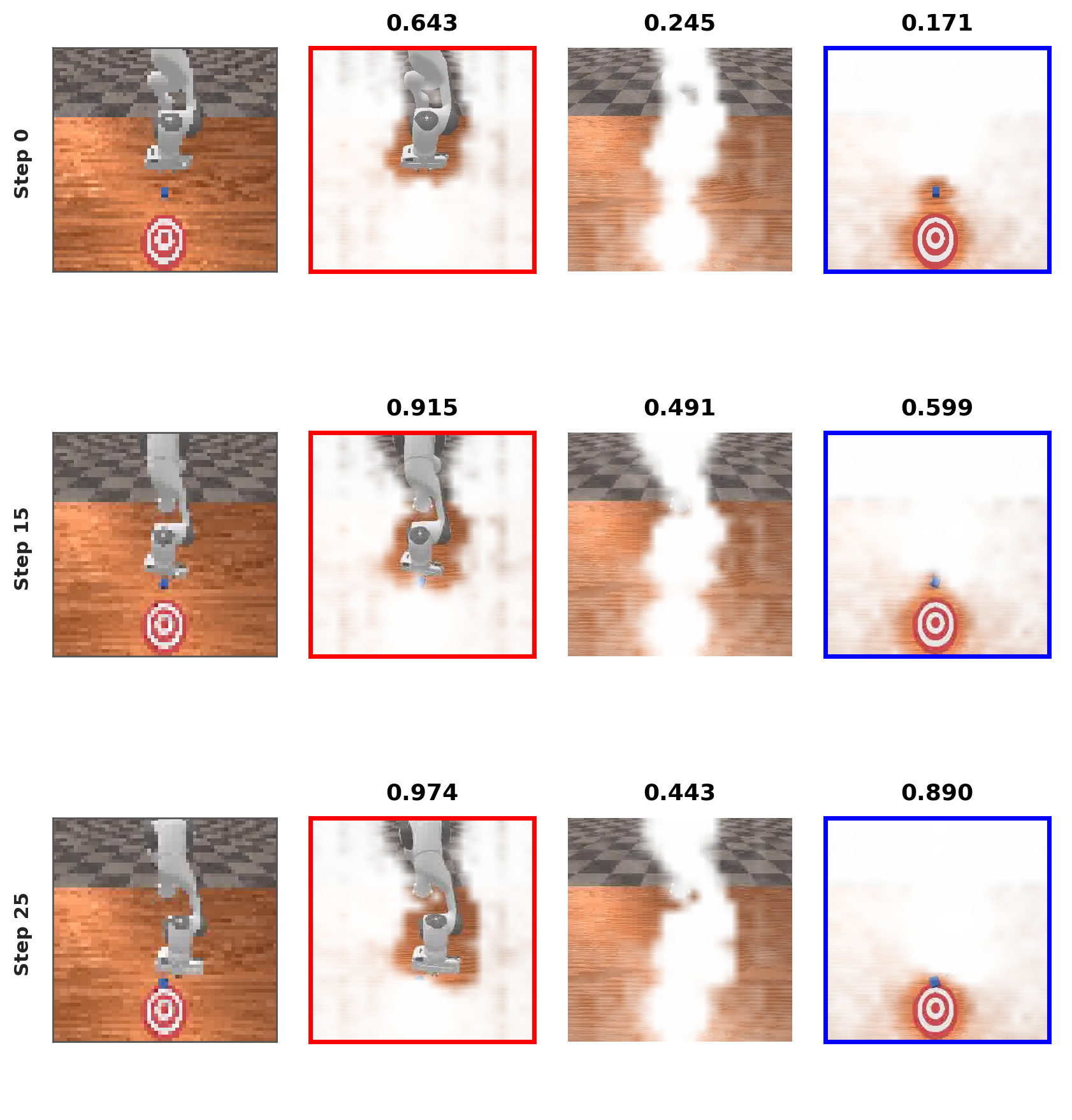}
\caption{Per-slot causality scores for the policy transformer in the Cube Pushing Task. Each row corresponds to a time step and shows the real observation from the environment together with attention maps produced by the Slot Contrast model for each slot inferred by the Slot Contrast model. The number above each slot denotes its causality score $\alpha_t^i$, indicating the probability that the corresponding object is causally relevant for policy prediction. Red bounding boxes indicate agent-related objects,
blue bounding boxes indicate target objects. The causality score is highest for the target and agent objects across most time steps.} 
\end{figure}

\begin{figure}[H] 
\includegraphics[width=\textwidth]{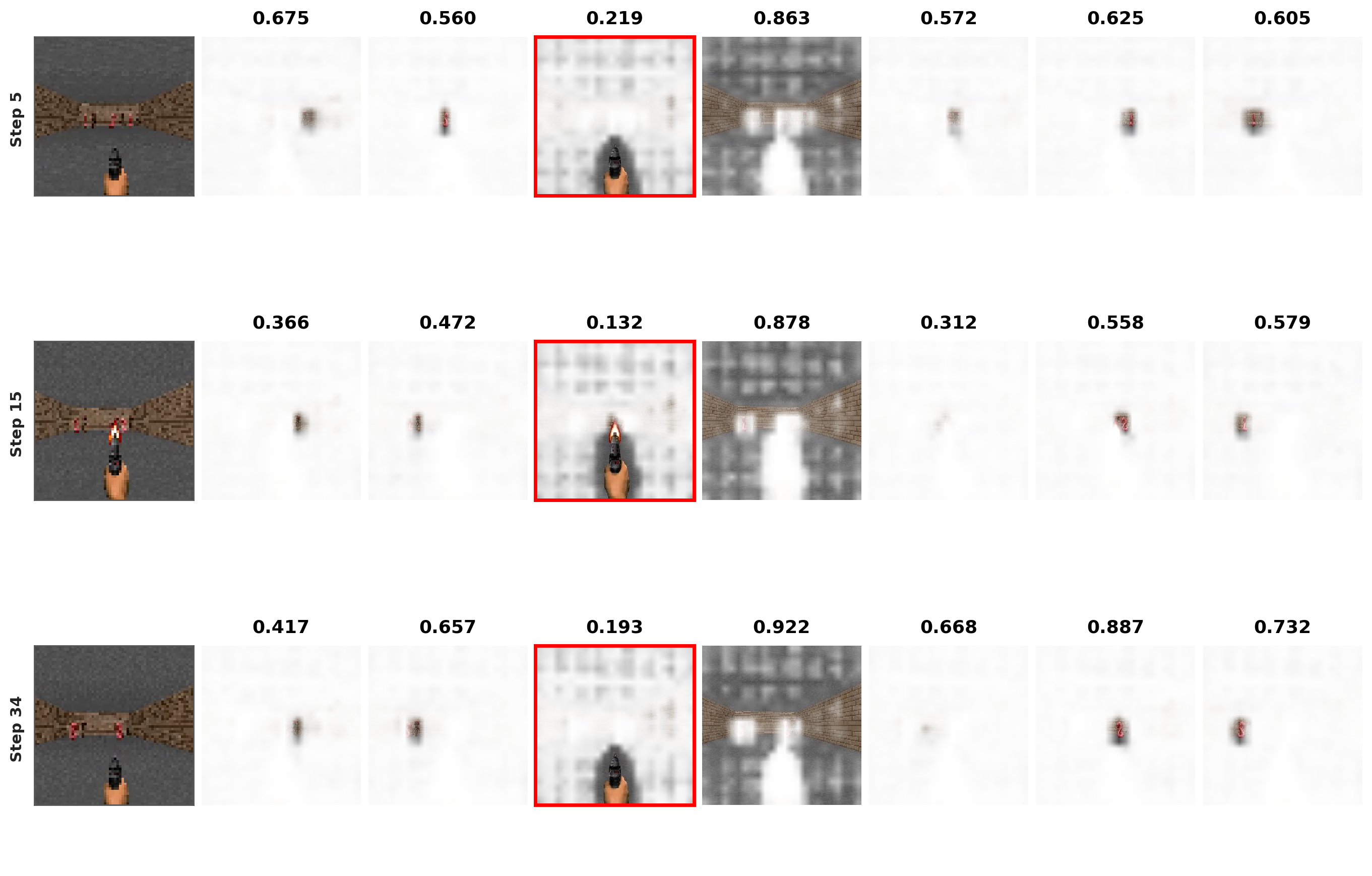}
\caption{
Per-slot causality scores for the value transformer in the Defend The Line Task. Each row corresponds to a time step and shows the real observation from the environment together with attention maps produced by the Slot Contrast model for each slot inferred by the Slot Contrast model. The number above each slot denotes its causality score $\alpha_t^i$, indicating the probability that the corresponding object is causally relevant for value prediction. Red bounding boxes indicate agent-related object. All objects, including the agent and the monsters, receive high causality scores. However, the model incorrectly assigns a high causality score to a background object.
}
\end{figure}

\begin{figure}[H] 
\includegraphics[width=\textwidth]{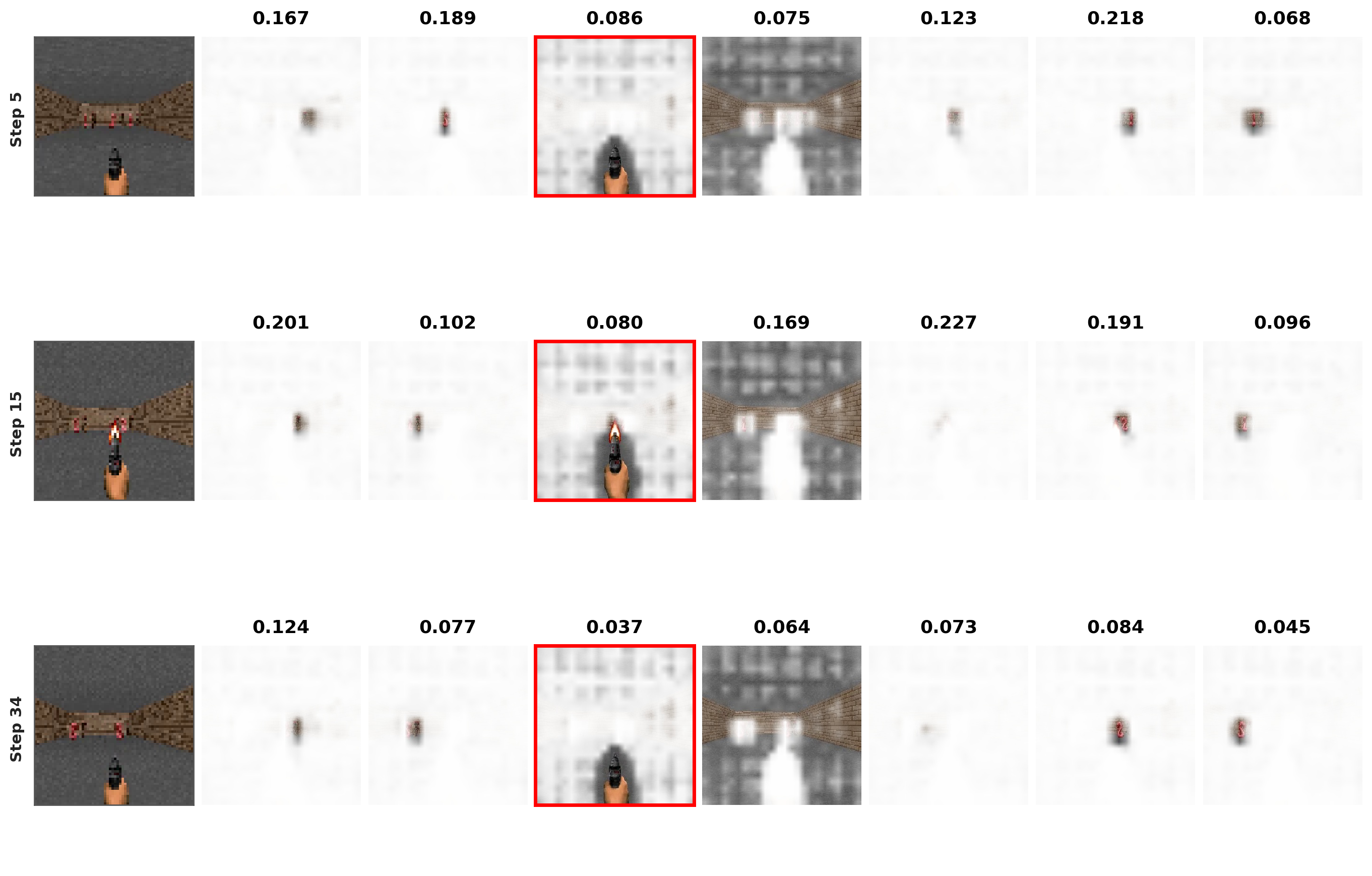}
\caption{Per-slot causality scores for the policy transformer in the Defend The Line Task. Each row corresponds to a time step and shows the real observation from the environment together with attention maps produced by the Slot Contrast model for each slot inferred by the Slot Contrast model. The number above each slot denotes its causality score $\alpha_t^i$, indicating the probability that the corresponding object is causally relevant for policy prediction. Red bounding boxes indicate agent-related object. All objects, including the agent and the monsters, receive high causality scores. However, the model incorrectly assigns a high causality score to a background object.} 
\end{figure}

\section{SAVI`s visualizations for SOLD}
\label{app:savi}

\begin{figure}[H] 
\includegraphics[width=\textwidth]{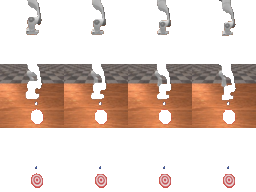}
\caption{Hard attention maps produced by the SAVI model in the Cube Pushing task.} 
\end{figure}

\begin{figure}[H] 
\includegraphics[width=\textwidth]{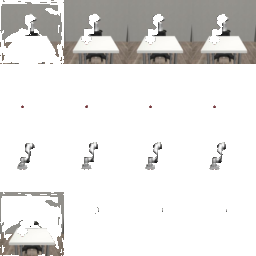}
\caption{Hard attention maps produced by the SAVI model in the Block Lifting task.} 
\end{figure}

\begin{figure}[H] 
\includegraphics[width=\textwidth]{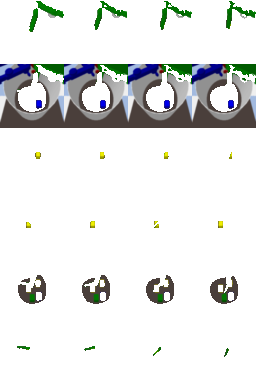}
\caption{Hard attention maps produced by the SAVI model in the Object Reaching task.} 
\end{figure}



\end{document}